\documentclass[pre,singlecolumn,notitlepage,superscriptaddress,longbibliography]{revtex4-1}
\usepackage[utf8]{inputenc}
\usepackage{bm}
\usepackage{amsmath}
\usepackage{amssymb}
\usepackage{hyperref}
\usepackage{graphicx}
\usepackage{bm}
\usepackage{subfig}
\usepackage{xcolor}
\newcommand{\erf}{\text{erf}}
\newcommand{\erfc}{\text{erfc}}
\usepackage[export]{adjustbox}
\begin{document}

\title{Expectation propagation on the diluted Bayesian classifier}

\author{Alfredo Braunstein}
\affiliation{Department of Applied Science and Technologies (DISAT), Politecnico di Torino, Corso Duca Degli Abruzzi 24, Torino, Italy}
\affiliation{Italian Institute for Genomic Medicine, IRCCS Candiolo, SP-142, I-10060 Candiolo (TO) - Italy}
\affiliation{INFN Sezione di Torino, Via P. Giuria 1, I-10125 Torino, Italy}
\author{Thomas Gueudr\'{e}}
\altaffiliation[Currently at ]{Amazon Alexa, Torino}
\affiliation{Department of Applied Science and Technologies (DISAT), Politecnico di Torino, Corso Duca Degli Abruzzi 24, Torino, Italy}

\author{Andrea Pagnani}
\affiliation{Department of Applied Science and Technologies (DISAT), Politecnico di Torino, Corso Duca Degli Abruzzi 24, Torino, Italy}
\affiliation{Italian Institute for Genomic Medicine, IRCCS Candiolo, SP-142, I-10060 Candiolo (TO) - Italy}
\affiliation{INFN Sezione di Torino, Via P. Giuria 1, I-10125 Torino, Italy}

\author{Mirko Pieropan}
\email[Author to whom any correspondence should be addressed. Email: ]{mirko.pieropan@polito.it}
\affiliation{Department of Applied Science and Technologies (DISAT), Politecnico di Torino, Corso Duca Degli Abruzzi 24, Torino, Italy}

\begin{abstract}
Efficient feature selection from high-dimensional datasets is a very important challenge in many data-driven fields of science and engineering.
We introduce a statistical mechanics inspired strategy that addresses the problem of sparse feature selection in the context of binary classification by leveraging a computational scheme known as expectation propagation (EP). The algorithm is used in order to train a continuous-weights perceptron learning a classification rule from a set of (possibly partly mislabeled) examples provided by a teacher perceptron with diluted continuous weights. 
We test the method in the Bayes optimal setting under a variety of conditions and compare it to other state-of-the-art algorithms based on message passing and on expectation maximization approximate inference schemes. Overall, our simulations show that EP is a robust and competitive algorithm in terms of variable selection properties, estimation accuracy and computational complexity, especially when the student perceptron is trained from correlated patterns that prevent other iterative methods from converging. Furthermore, our numerical tests demonstrate that the algorithm is capable of learning online the unknown values of prior parameters, such as the dilution level of the weights of the teacher perceptron and the fraction of mislabeled examples, quite accurately. This is achieved by means of a simple maximum likelihood strategy that consists in minimizing the free energy associated with the EP algorithm.\end{abstract}

\maketitle

\section{Introduction}

The problem of extracting sparse information from high dimensional
data is among the most interesting challenges in theoretical computer
science with many applications ranging from computational biology to
combinatorial chemistry, neuroscience and natural language processing
\cite{guyon2008, hastie2009}. As a specific example, next
generation sequencing and, in general, the ongoing technological
revolution related to high-throughput technologies in biology pose
very stringent requirements to the algorithmic techniques that are
supposed to analyze the data that are produced and made publicly
available through easily accessible databases. Just to give some orders
of magnitude, a typical genetic screening for cancer-related
pathologies -- freely available from The Cancer Genome Atlas web site
\cite{TCGA} -- involves measurement of activity or genetic sequence
variation over $\sim 23,000$ genes measured on patient cohorts that
typically count around 1,000 individuals divided into cases and
controls (lung and colorectal cancer are an exception, with $\sim
10,000$ individuals screened in each dataset). Here, a typical task
is to determine the genotypic signature related to the disease that
typically involves $O(10^2)$ genes from 23,000 measured
probes. Such problem can be simply formulated in terms of the
following classification problem: given the activity and/or the
genetic alterations of an individual, find a simple rule involving a
small -- possibly the smallest -- subset of genes to assess the
probability for the individual to develop the disease. There are two
main difficulties in this task: (i) typically genes
act in a combinatorial and non linear manner, (ii) individual samples
turn out to be statistically very correlated.

Historically, the problem of sparse feature selection in
classification tasks has been divided into two complementary
computational methods \cite{hastie2009}: (i) {\em wrappers} that
exploit the learning mechanism to produce a prediction value related
score for the sought signature, (ii) {\em filters} where the signature
extraction is a data pre-processing, typically unrelated to the
classification task.

From the point of view of information theory, the problem of sparse
feature selection in classification is strictly related to Compressive
Sensing (CS), one of the most studied methods for data acquisition,
with interesting applications in several other research fields
\cite{candes2008,baraniuk2011}. CS was originally proposed as a new
low-rate signal acquisition technique for compressible signals
\cite{donoho2006, candes2006, candes2008} and is formulated as
follows: given $M<N$, a vector $\bm z\in {\mathbb R}^M$ and a linear
operator of maximal rank $\mathbf{X}\in{\mathbb R}^{M\times N}$ often
referred to as the \emph{measurement} or \emph{sensing} matrix, the CS
problem consists in determining the unknown sparse vector $\bm w \in
{\mathbb R}^N$ that is linked to its compressed projection $\bm z$ by
means of the linear transformation $\bm z =\mathbf{X} {\bm w}$, where
$\mathbf{X}$ and $\bm z$ are assumed to be known. Although research in
CS has still many open challenges to face, very stringent results are
known about the general conditions for the existence and uniqueness of
the solution. Among the different algorithms that have been proposed
in order to reconstruct efficiently the signal, many use techniques
borrowed from the statistical mechanics of disordered systems \cite{
  donoho2009, kabashima2009,ganguli2010, krzakala2010}.

More recently, the so called 1-bit CS (1BCS) has been proposed as a
strategy to deal with the problem of inferring a sparse signal knowing
only the sign of the data of the linear measurements: ${\boldsymbol \sigma} =
\mathrm{sign} (\mathbf{X} {\bm w})$, where $\mathrm{sign}({\bm z})$ is
a vector with elements $z_i/|z_i|$ for $z_i \neq 0$. Besides being of
interest for signal transmission related problems where discarding the
amplitude of the signal can significantly reduce the amount of
information to be stored/relayed \cite{boufounos2008, lee2012}, this
problem can also be interpreted in terms of sparse boolean
classification tasks. The most widely adopted inference scheme in CS
is the so-called LASSO regression or $L_1$-norm minimization
\cite{tibshirani1996}, as originally proposed in the context of 1BCS
in \cite{boufounos2008}. However, it is clear that the most efficient
solution from the point of view of optimal dilution of the problem
should be achieved by a $L_0$-pseudonorm, where non-zero parameters
are indeed penalized independently of their non-zero
value. Unfortunately, dealing with the non-convex $L_0$-regularization
is not so simple as it typically leads to phase transitions that make
the problem computationally intractable. A practical solution to the
problem is to restrict the space of parameters to a discrete set,
where effectively the $L_0$-pseudonorm is equivalent to the
more amenable $L_1$ case \cite{braunstein2008, braunstein2008b,
  pagnani2009, bailly2010, molinelli2013}. As far as continuous
parameters are concerned, different strategies have been
proposed. First, from the statistical physics community side, an
approach pursuing this direction consists in a perceptron whose
continuous parameters are masked by boolean variables mimicking
dilution \cite{kabashima2003, uda2006, kabashima2008,
  xu2013}. Attempts to characterize theoretically the phase space
diagram and the structure of the transition through the replica method
have been reported in
\cite{lage2009,lage_allerton,lage2013}. Variations of the generalized
approximate message passing technique (GAMP) were employed in
\cite{onebitAMP}, as it provides a tractable and efficient way to
perform minimum mean squared error (MMSE) estimation on the variables
to be retrieved when the matrix of patterns is large and Gaussian
i.i.d.. However, for more general pattern matrices, GAMP convergence
is not guaranteed, which has led to the extension of algorithms of the
vector AMP (VAMP) type \cite{VAMP} to generalized linear models
\cite{schniter2016vector,grVAMP}, including perceptron learning.

On the computer science side, many other algorithms for 1BCS combining
the enforcement of sparsity and of sign consistency constraints were
also proposed, building upon analogous algorithms developed for
standard CS.  Examples of methods for error-free sparse signal
retrieval from one-bit quantized measurements include greedy
approaches which iteratively determine the most appropriate sparse
support given the sign measurements, such as matching sign pursuit
\cite{MSP}, as well as binary iterative hard thresholding (BIHT)
\cite{BIHT}, where an $L_1$-based convex consistency-enforcing
objective function minimization is alternated with a thresholding
operation that selects the $K$ largest elements.  The problem of noisy
1BCS was addressed, for instance, in \cite{AOP,NARFPI,R1BCS}.
However, among these examples, only \cite{R1BCS} proposes an
algorithm which does not require the prior knowledge of the number of
corrupted sign measurements. Here, the one-bit measurement errors are
modeled by introducing a sparse vector $\boldsymbol{s}$ whose nonzero
components produce the sign mismatches as ${\boldsymbol \sigma} = \mathrm{sign}
(\mathbf{X} {\bm w}+\boldsymbol{s})$. The algorithm attempts to
identify the sign errors and to retrieve the sparse signal $\bm w$
using a variational expectation-maximization (EM) based inference scheme.

\begin{figure}
\centering
\begin{subfloat}[]{
\includegraphics[width=0.2\textwidth]{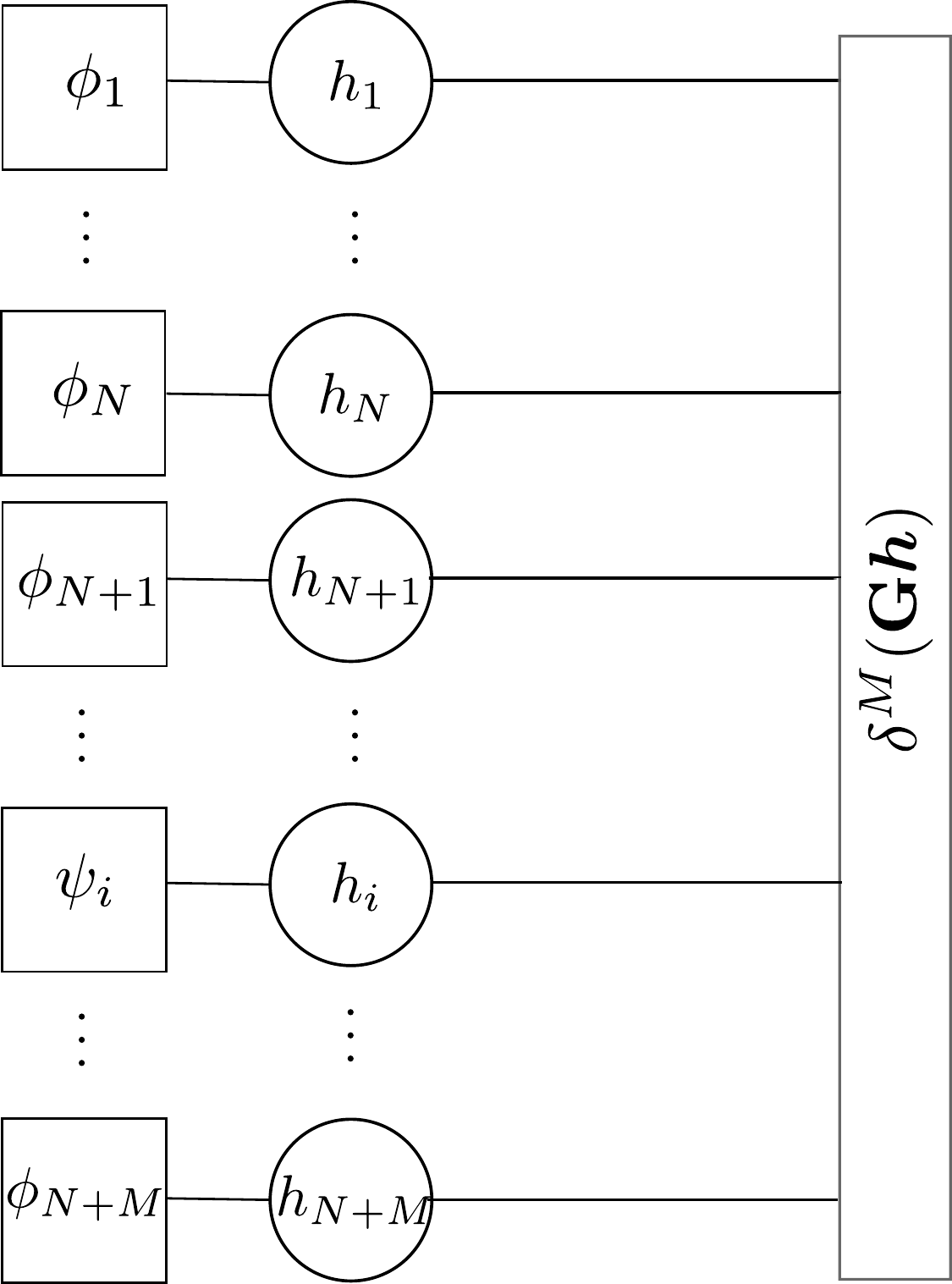}
\label{fig:EP-FG}}
\end{subfloat}
\qquad\qquad
\begin{subfloat}[]{
\includegraphics[width=0.4\textwidth]{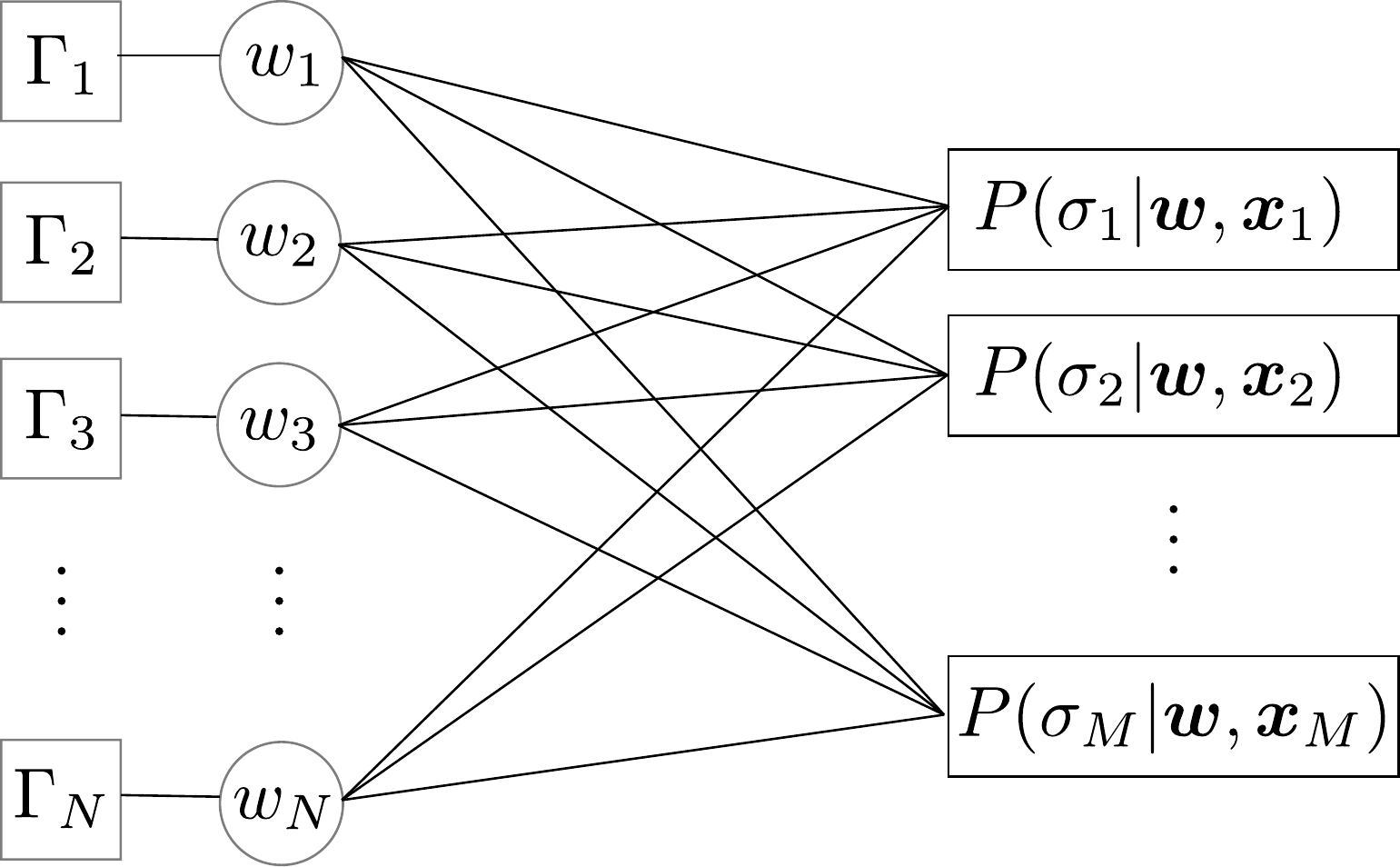}
\label{fig:1bitAMP-FG}}
\end{subfloat}
\\
\begin{subfloat}[]{
\includegraphics[width=0.7\textwidth]{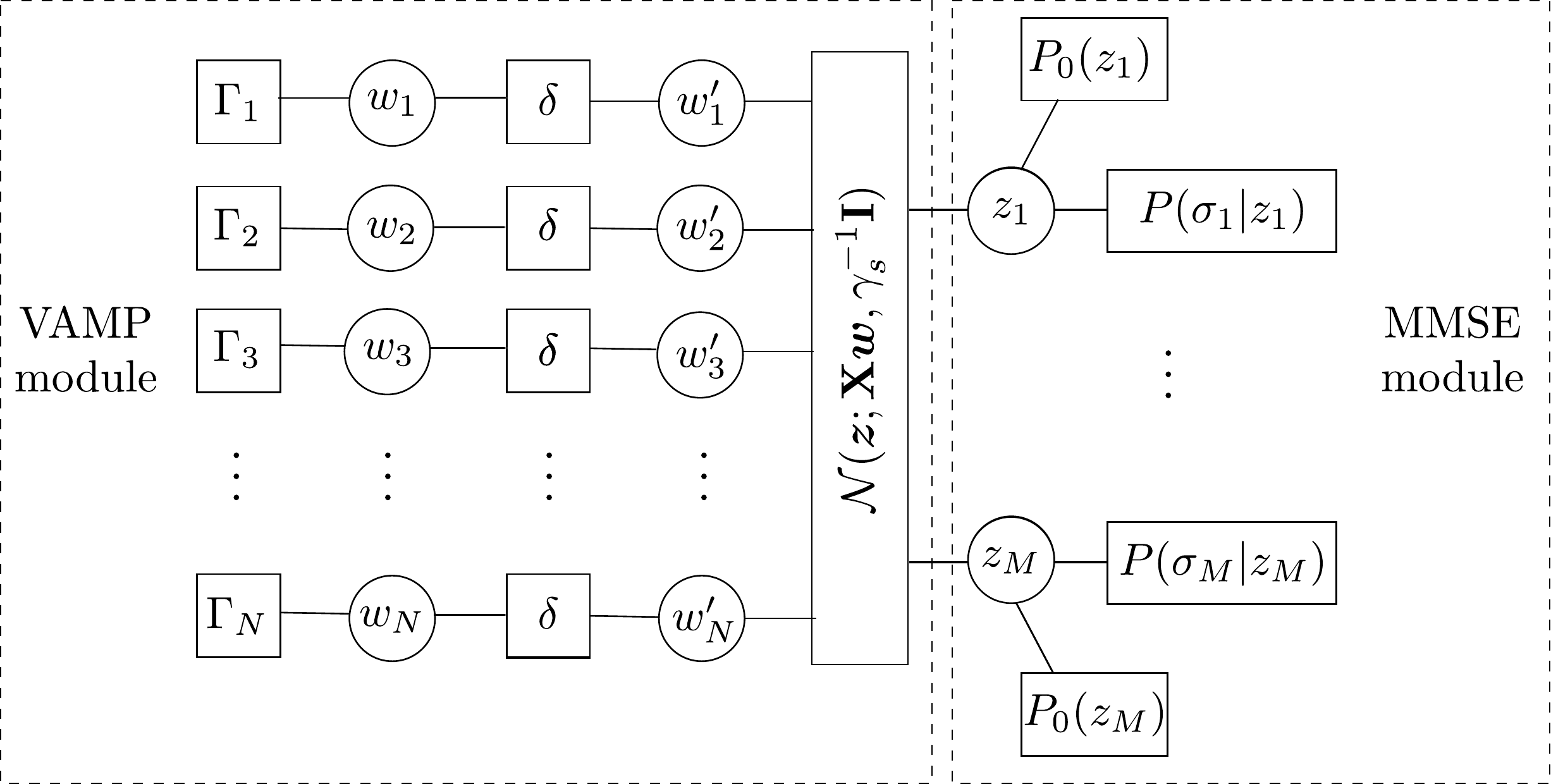}
\label{fig:grVAMP-FG}}
\end{subfloat}
\\
\begin{subfloat}[]{
\includegraphics[width=0.85\textwidth]{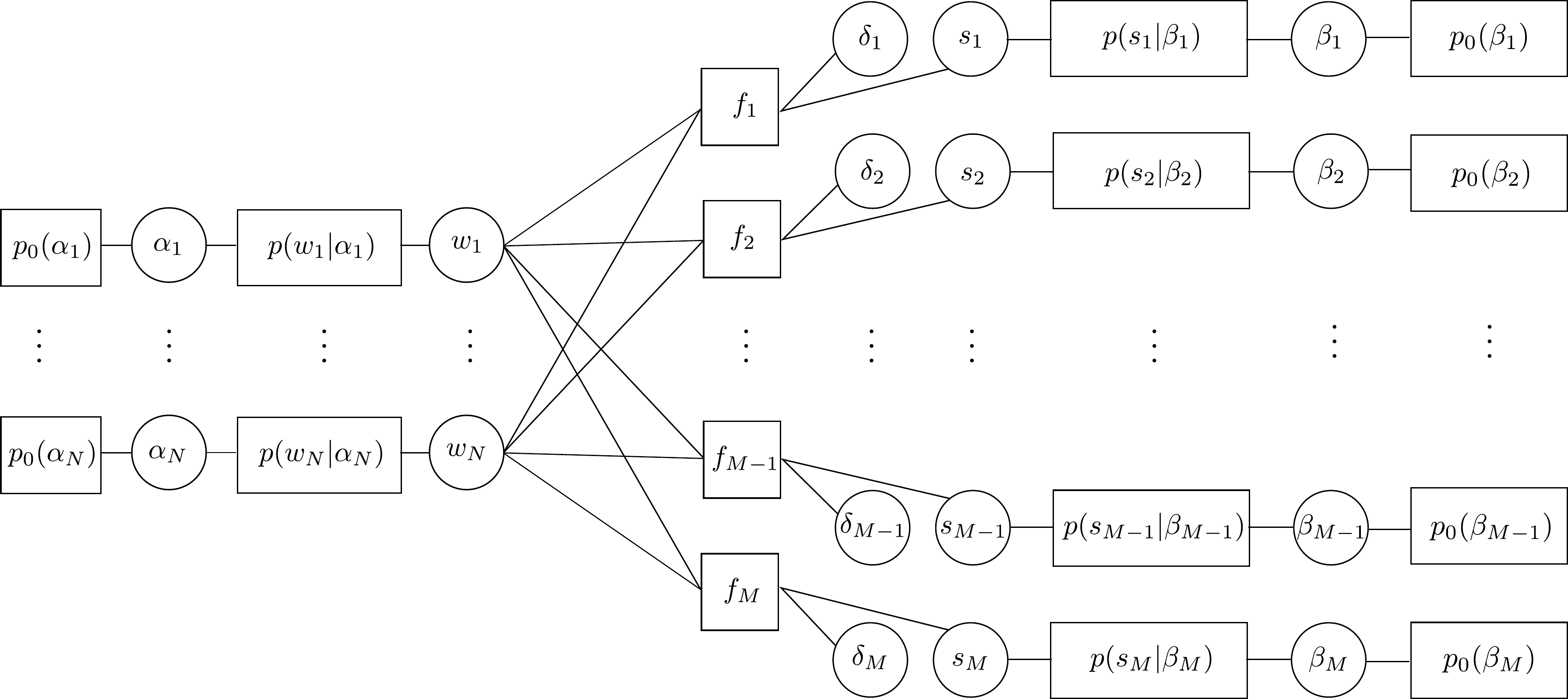}
\label{fig:R1BCS-FG}}
\end{subfloat}
\caption{Factor graphs associated with (a) EP, (b) AMP, (c) grVAMP and (d) R1BCS. Variable nodes are represented as circles and function nodes are represented as squares and the notation has been made consistent with that employed in this paper. 
(a) Tilted distributions in EP (see Section \ref{subsec:EP} for more details), where $\psi$ denotes the exact prior and $\phi$ the approximated Gaussian prior factors in the EP approximation. (b) Factor graph related to AMP. In 1bitAMP, $\Gamma$ are spike-and-slab priors and $P(\sigma_\mu|\boldsymbol{w},\boldsymbol{x}_\mu)=\Theta(\sigma_\mu\boldsymbol{x}^T_\mu\boldsymbol{w})$. (c) Factor graph related to the grVAMP approximation, where we have emphasized the VAMP and MMSE modules composing the algorithm, $\Gamma$ are the same as in (b) and $P(\sigma_\mu|z_\mu)=\Theta(\sigma_\mu z_\mu)$. In (b) and (c), $\Theta$ denotes the Heaviside theta function. (d) Factor graph of the function appearing in the lower bound maximized in R1BCS. The distributions $p(w_i|\alpha_i)p(\alpha_i)$ and $p(s_\mu|\beta_\mu)p(\beta_\mu)$ are hierarchical Gaussian-inverse-Gamma priors assigned to the weights and to the noise components $s_\mu$, respectively, appearing in ${\boldsymbol \sigma} = \mathrm{sign}
(\mathbf{X} {\bm w}+\boldsymbol{s})$. The quantities $\alpha_i$ and $\beta_\mu$ are hyperparameters, whereas the quantities $\delta_\mu$ are variational parameters optimized in the maximization step of the algorithm.} 
\label{fig:factor-graphs}
\end{figure}

In this work we propose a new {\em wrapper} strategy where both the
variable selection and the classification tasks are simultaneously
performed through Expectation Propagation (EP), an iterative scheme to
approximate intractable distributions that was introduced first in the
field of statistical physics
\cite{opper_gaussian_2000,opper_adaptive_2001} and shortly after in
the field of theoretical computer science
\cite{minka_expectation_2001}.  In analogy to what presented in
\cite{braunstein2017} in the context of sampling the space of high
dimensional polytopes, we show that, by approximating the
computationally intractable posterior distribution $P(\bm w |
\boldsymbol \sigma, \bf X)$ through a tractable multivariate
probability density $Q(\bm w | \boldsymbol \sigma, \bf X)$, we are
able to solve both efficiently and accurately the 1BCS problem. We
compare our results to those obtained from the AMP and VAMP based
schemes proposed in Refs. \cite{onebitAMP} and \cite{grVAMP},
respectively, and to those given by the EM based approach of
Ref. \cite{R1BCS}. We provide the factor graphs associated with these
algorithms in Fig. \ref{fig:factor-graphs}.  We show through simulations that one of the main strengths of the EP-based approach is that it is effective on a wider family of measurement matrices with respect to other relatively similar algorithms such as VAMP and AMP. 

The paper has the following structure: after this introduction, in
Sec.~\ref{sec:methods} we define the problem, and introduce the EP
algorithm. In Sec.~\ref{sec:results} we present extensive numerical
simulations both in the noiseless and noisy case. Here both i.i.d. and
correlated measurement matrices are analyzed. Finally, in
Sec.~\ref{sec:conclusions} we summarize the results of the paper 
and draw the conclusions.

\section{Methods}
\label{sec:methods}
\subsection{The diluted perceptron as a linear estimation problem and its statistical mechanics setup}

We consider a student perceptron with $N$ input units and continuous
weights $\boldsymbol{w}\in\mathbb{R}^N$. We assume that the
connections are diluted and that only a fraction $\rho$ of them are
nonzero. We also assume that $M$ real-valued patterns
$\boldsymbol{x}_{\tau}\in\mathbb{R}^N$ are presented to the perceptron
and that a binary label $\sigma_{\tau}$, $\tau=1,\dots,M$ has already
been assigned to each of them as a result of the classification
performed by a teacher perceptron with sparse continuous weights
$\boldsymbol{B}$. The task of the student perceptron is to learn the
input/output association based on the examples
$(\boldsymbol{x}_{\tau},\sigma_{\tau}),\ \tau=1,\dots, M$ provided by
the teacher:
\begin{equation} 
  \sigma_{\tau}=\text{sign}(\bm{w}^{T}\bm{x}_\tau),\quad \tau=1,\dots,M,
  \label{eq:classification} 
\end{equation}
where we use the convention that $\text{sign}(0)=1$. For each example
$\tau$, the rule \eqref{eq:classification} is equivalent to the condition:
\begin{equation}
  \left(\sigma_{\tau}\bm{x}_\tau^{T}\right)\bm{w}\geq0\,.
  \label{eq:consistency_constraint}
\end{equation}
We now introduce the auxiliary variables,
$y_{\tau}:=\left(\sigma_{\tau}\bm{x}^{T}_{\tau}\right)\bm{w}$, and the data
matrix:
\begin{equation}
  \mathbf{X_\sigma}=\left(\begin{array}{c}
    \sigma_{1}\bm{x}^{T}_{1}\\
    \sigma_{2}\bm{x}^{T}_{2}\\
    \vdots\\
    \sigma_{M}\bm{x}^{T}_{M}
  \end{array}\right)\,.
\end{equation}
Through the previous definitions, we can define the following linear estimation
problem:
\begin{equation}
  \bm{y}=\mathbf{X}_\sigma\bm{w},
  \label{eq:perceptronLEP}
\end{equation}
where the variables to be inferred are both $\bm{y}$ and $\bm{w}$. As
we will show below, the positivity constraints in
Eq.~\eqref{eq:consistency_constraint} will be enforced in terms of a
prior distribution on the $\bm{y}$ variables.

The linear estimation problem expressed in Eq.
\eqref{eq:perceptronLEP} can be addressed in a Bayesian setting: by
introducing the variable vector
$\bm{h}=\left(w_{1},\dots,w_{N},y_{1},\dots,y_{M}\right)^T$ and the
energy function:
\begin{equation}
E(\bm{w},\bm{y})=\Vert\bm{y}-\mathbf{X}_\sigma\bm{w}\Vert^{2}=\bm{h}^{T}\mathbf{E^{-1}}\bm{h},\quad\mathbf{E}^{-1}=\left(\begin{array}{cc}
  \mathbf{X}_\sigma^{T}\mathbf{X}_\sigma &
  -\mathbf{X}_\sigma^T\\ -\mathbf{X}_\sigma & \mathbf{I}
\end{array}\right),
\end{equation}
the likelihood of the set of $N$ weights of the perceptron can be
expressed as the Boltzmann distribution associated with
$E(\boldsymbol{w},\boldsymbol{y})$, which reads:
\begin{equation}
    \mathcal{L}(\boldsymbol{w})=P(\sigma_1,\dots,\sigma_M|\boldsymbol{w})=\frac{1}{Z}e^{-\beta E(\boldsymbol{w},\boldsymbol{y})},
\end{equation}
where, from a statistical physics standpoint, $\beta$ plays the role
of an inverse temperature. In the absence of noise, it is convenient
to consider the zero temperature limit of this likelihood
$\mathcal{L}(\boldsymbol{w})\xrightarrow{\beta\rightarrow\infty}{\delta(\boldsymbol{y}-\mathbf{X}_\sigma\boldsymbol{w})}$,
where $\delta(x)$ denotes the Dirac delta distribution.

We also introduce prior distributions in order to encode the constraints to which the variables $w_i$, $i=1,\dots,N$, and $y_\tau$, $\tau=1,\dots,M$,  are subject. The sparsity assumption on the weights $\boldsymbol{w}$ are expressed in terms of a spike-and-slab prior \cite{spike-and-slab}:
\begin{equation}
\Gamma(w_{i})=(1-\rho)\delta(w_{i})+\rho\sqrt{\frac{\lambda}{2\pi}}e^{-\frac{\lambda w_{i}^{2}}{2}},\quad i=1,\dots,N.
\label{eq:spike-and-slab}
\end{equation}
If the labels of the teacher are not corrupted by noise, then the auxiliary variables $\boldsymbol{y}$ need to fulfill the positivity constraint \eqref{eq:consistency_constraint}, which can be expressed in terms of the pseudoprior:
\begin{equation}
    \Lambda(y_{\tau})=\Theta(y_{\tau}),\quad\tau=1,\dots,M.
\end{equation}
On the other hand, if noise at the output of the teacher perceptron is present, one may assume that the labels provided by the teacher perceptron are assigned according to the following process \cite{nishimori2001}:
\begin{equation}
    \tilde{\sigma}=\begin{cases}
\text{sign}(\boldsymbol{B}^T\boldsymbol{x}) & \text{with probability}\quad \eta\\
-\text{sign}(\boldsymbol{B}^T\boldsymbol{x}) & \text{with probability}\quad 1-\eta
\end{cases}
\end{equation}
and that the student receives the altered examples $(\boldsymbol{x}_\mu,\tilde{\sigma}_\mu),\ \mu=1,\dots,M$. In this case, if the process that flips the labels is known, then it may be encoded in the pseudoprior $\Lambda$ as follows:
\begin{equation}
    \Lambda(y_\mu)=\eta\Theta(y_\mu)+(1-\eta)\Theta(-y_\mu).
\end{equation}
In general, the parameters $\rho$, $\lambda$ and $\eta$ are not known
and need to be learned by the student perceptron in the training
phase. Finally, by Bayes' rule, the posterior distribution of both
weights and auxiliary variables read:
\begin{equation}
    P(\boldsymbol{w},\boldsymbol{y})=\frac{1}{Z_P}\delta(\boldsymbol{y}-\mathbf{X}_\sigma\boldsymbol{w})\prod_{i=1}^{N}\Gamma_i(w_i)\prod_{\tau=1}^{M}\Lambda_\tau(y_\tau).
    \label{eq:posterior}
\end{equation}

\subsection{Learning the weights via expectation propagation}
\label{subsec:EP}
\subsubsection{Zero temperature formulation}
We wish to infer the values of the weights by estimating the
expectation values of the marginals of the distribution
\eqref{eq:posterior}, as this strategy minimizes the associated mean
squared error. However, the latter marginalizations are intractable
and we need to resort to approximation methods. Here we propose an
expectation propagation scheme based on the zero temperature
formulation presented in \cite{csep2020} in order to solve the
problem.

Starting from the linear system $\mathbf{X}_\sigma\bm{w}=\bm{y}$, we
notice that it can be written as the homogeneous system:
\begin{equation}
    \mathbf{G}\bm{h}=\bm{0},
\end{equation}
where $\mathbf{G}=\left(-\mathbf{X}_\sigma|\mathbf{I}\right)$ and
$\mathbf{I}$ is the $M\times M$ identity matrix.

The intractable posterior distribution reads:
\begin{equation}
    P(\boldsymbol h)= \frac{1}{Z_P}\delta^M(\mathbf{G}\bm{h})\prod_{i\in W}\Gamma_i(h_i)\prod_{\tau\in Y}\Lambda_\tau(h_\tau),
    \label{eq:posterior_zerotemp}
\end{equation}
where $W=\{1,\dots,N\}$, $Y=\{N+1,\dots,N+M\}$ and $\delta^M(\boldsymbol z)$ denotes the $M$-dimensional Dirac delta distribution.
We introduce Gaussian approximating factors:
\begin{equation}
\phi_i(h_i)=\exp\left(-\frac{(h_i-a_i)^2}{2d_i}\right),
\label{eq:gaussianfactors}
\end{equation}
and a fully Gaussian approximation of the posterior distribution \eqref{eq:posterior_zerotemp}, in which all priors $\Gamma$ and $\Lambda$ are replaced by factors of the form \eqref{eq:gaussianfactors}:
\begin{equation}
    Q(\boldsymbol{h})=\frac{1}{Z_Q}\delta^M(\mathbf{G}\bm{h})\prod_{i\in W}\phi(h_i;a_i,d_i)\prod_{\tau\in Y}\phi(h_\tau;a_\tau,d_\tau).
\end{equation}
$Q(\bm{h})$ can be equivalently expressed as:
\begin{equation}
    Q(\boldsymbol{h})=\frac{1}{Z_Q}\delta^M(\mathbf{G}\bm{h})\exp\left(-\frac{1}{2} (\bm{w} - \bar{\bm{w}})^T \bm \Sigma _W^{-1} (\bm{w} - \bar{\bm{w}})\right),
    \label{eq:Q_distr}
\end{equation}
where the covariance matrix $\bm \Sigma _W$ and the mean $\bar{\bm w}$ in Eq. \eqref{eq:Q_distr} are given, respectively, by:
\begin{equation}
    \boldsymbol\Sigma _W^{-1}=\sum_{i\in W}\frac{1}{d_i}\boldsymbol{e}_i \boldsymbol{e}^T_i+\mathbf{X}_\sigma^T \left( \sum_{i\in Y}\frac{1}{d_i}\boldsymbol{e}_i \boldsymbol{e}^T_i \right)\mathbf X_\sigma,
    \label{eq:cov_Q}
\end{equation}
and by:
\begin{equation}
\bar{\bm{w}}= \boldsymbol\Sigma _W\left(\sum_{i\in W}\frac{a_i}{d_i}\boldsymbol{e}_i+\sum_{i\in Y}\frac{a_i}{d_i}\mathbf{X}_\sigma^T \boldsymbol{e}_i \right).
\end{equation}
Here, $\bm{e}_i$ denotes the $i$-th basis vector of the standard basis of $\mathbb{R}^{N}$ (resp. $\mathbb{R}^{M}$) if $i \in W$ (resp. $i \in Y$).
Notice that the marginal distributions of $Q(\boldsymbol h)$ for each variable $h_i$ are also Gaussian, with means $\bar{h}_i$ given by:
\begin{equation}
\bar{h}_i=
\begin{cases}
\bar{w}_i, \quad i\in W
\\
{\boldsymbol e}_i^T\mathbf{X}_{\sigma} \bar{\boldsymbol{w}}, \quad i\in Y,
\end{cases}
\end{equation}
and variances $\Sigma_{ii}$ given by:
\begin{equation}
\Sigma_{ii}=
\begin{cases}
\boldsymbol{e} _i^T \boldsymbol{\Sigma}_W \boldsymbol{e} _i, \quad i\in W,
\\
\left( \boldsymbol{e} _i^T \mathbf{X}_\sigma \right) \bm{\Sigma}_W \left(\mathbf{X}_\sigma^{T} \boldsymbol{e} _i \right), \quad i\in Y,
\end{cases}
\end{equation}
where for $i\in Y$ we took advantage of the linear constraints $\bm{y}=\mathbf{X}_\sigma\bm{w}$.
Notice that the full $(N+M)\times(N+M)$ covariance matrix $\Sigma$ (whose diagonal entries are defined in the previous equation) reads:
\begin{equation}
\boldsymbol \Sigma = \left(
\begin{array}{cc}
\boldsymbol \Sigma _W & \Sigma _W \mathbf{X}_\sigma ^T \\ 
 \mathbf{X}_\sigma  \Sigma _W & \mathbf{X}_\sigma  \Sigma _W \mathbf{X}_\sigma ^T
\end{array}
\right).
\end{equation}

We now introduce $N+M$ tilted distributions $Q^{(i)}(\boldsymbol{h})$ for $i=1,\dots,N+M$.
In particular, if $i\in W$, we have:
\begin{equation}
    Q^{(i)}(\boldsymbol{h})=\frac{1}{Z_{Q^{(i)}}}\delta^M(\mathbf{G}\bm{h})\Gamma_i(h_i)\prod_{i\in W\setminus\{i\}}\phi(h_i;a_i,d_i)\prod_{\tau\in Y}\phi(h_\tau;a_\tau,d_\tau),
    \label{eq:leave-one-out-Gamma}
\end{equation}
whereas, if $i\in Y$:
\begin{equation}
    Q^{(i)}(\boldsymbol{h})=\frac{1}{Z_{Q^{(i)}}}\delta^M(\mathbf{G}\bm{h})\Lambda_i(h_i)\prod_{i\in W}\phi(h_i;a_i,d_i)\prod_{\tau\in Y\setminus\{i\}}\phi(h_\tau;a_\tau,d_\tau).
    \label{eq:leave-one-out-Lambda}
\end{equation}
The tilted distributions can be expressed as the product of one of the priors and a Gaussian cavity distribution:
\begin{equation}
    Q^{(i)}(\boldsymbol{h})=\psi_i(h_i)\Tilde{Q}^{(i)}(\bm{h}),
\end{equation}
where $\psi\in\{\Gamma,\Lambda\}$ and we have denoted the cavity distribution associated with the $i$-th variable by $\Tilde{Q}^{(i)}$:
\begin{equation}
    \Tilde{Q}^{(i)}(\bm{h}) = \frac{1}{Z_{Q^{(i)}}}\delta^M(\mathbf{G}\bm{h}) \exp\left(-\frac{1}{2} (\bm{w}-\bar{\bm{w}}^{(i)})^T \left(\bm{\Sigma}_W^{(i)}\right)^{-1} (\bm{w}-\bar{\bm{w}}^{(i)})\right).
    \label{eq:cavity-distr}
\end{equation}
A factor graph representation of the tilted approximation to the posterior distribution is given in Fig. \ref{fig:EP-FG}.

The cavity covariance matrices are given by the following expressions:
\begin{equation}
    \left(\bm{\Sigma}_W^{(i)}\right)^{-1} = \begin{cases} \sum_{j\in W\setminus\{i\}}\frac{1}{d_j}\boldsymbol{e}_j \boldsymbol{e}^T_j+\mathbf{X}_\sigma^T \left( \sum_{j\in Y}\frac{1}{d_j}\boldsymbol{e}_j \boldsymbol{e}^T_j \right)\mathbf X_\sigma, & \mbox{if } i \in W, \\ \sum_{j\in W}\frac{1}{d_j}\boldsymbol{e}_j \boldsymbol{e}^T_j+\mathbf{X}_\sigma^T \left( \sum_{j\in Y\setminus\{i\}}\frac{1}{d_j}\boldsymbol{e}_j \boldsymbol{e}^T_j \right)\mathbf X_\sigma, & \mbox{if } i \in Y. \end{cases}
\label{cavity_cov}
\end{equation}
whereas the cavity means read:
\begin{equation}
     \bar{\bm{w}}^{(i)} = \begin{cases} \boldsymbol\Sigma_W^{(i)}\left(\sum_{j\in W\setminus\{i\}}\frac{a_j}{d_j}\boldsymbol{e}_j+\sum_{j\in Y}\frac{a_j}{d_j}\mathbf{X}_\sigma^T \boldsymbol{e}_j \right), & \mbox{if } i \in W, 
     \\ 
     \boldsymbol\Sigma _W^{(i)}\left(\sum_{j\in W}\frac{a_j}{d_j}\boldsymbol{e}_j+\sum_{j\in Y\setminus\{i\}}\frac{a_j}{d_j}\mathbf{X}_\sigma^T \boldsymbol{e}_j \right), & \mbox{if } i \in Y. \end{cases}
\end{equation}

Similarly to what we obtained for the marginals of Eq. \eqref{eq:Q_distr}, we have that the marginals of Eq. \eqref{eq:cavity-distr} are Gaussian distributions with means:
\begin{equation}
\bar{h}^{(i)}_i=
\begin{cases}
\bar{w}^{(i)}_i, \quad \text{if } i\in W
\\
{\boldsymbol e}_i^T\mathbf{X}_{\sigma} \bar{\boldsymbol{w}}^{(i)}, \quad \text{if } i\in Y,
\end{cases}
\end{equation}
and variances:
\begin{equation}
\Sigma^{(i)}_{ii}=
\begin{cases}
\boldsymbol{e} _i^T \boldsymbol{\Sigma}^{(i)}_W \boldsymbol{e} _i, \quad \text{if } i\in W,
\\
\left( \boldsymbol{e} _i^T \mathbf{X}_\sigma \right) \bm{\Sigma}^{(i)}_W \left(\mathbf{X}_\sigma^{T} \boldsymbol{e} _i \right), \quad \text{if } i\in Y.
\end{cases}
\end{equation}

The yet to be determined means $\bm{a}$ and variances $\bm{d}$ of the Gaussian approximating factors \eqref{eq:gaussianfactors} are determined by minimizing the Kullback-Leibler divergence $D_{KL}(Q^{(i)}||Q)$ for all $i=1,\dots,N+M$. It can be shown that each of these minimizations is equivalent to matching the first and second moments of the tilted and of the fully Gaussian approximated distributions:
\begin{equation}
    \langle h_i \rangle_{Q^{(i)}}=\langle h_i \rangle_{Q},\quad \langle h^2_i \rangle_{Q^{(i)}}=\langle h^2_i \rangle_{Q}.
    \label{eq:mom_matching}
\end{equation}

The EP update equations follow from the moment matching conditions \eqref{eq:mom_matching}. In particular, recalling that the marginals of $Q(\bm{h})$ are Gaussian distributions, one can express $a_i$ and $d_i$ in terms of the means and variances of $Q^{(i)}$ and in terms of the means and variances of the associated tilted distributions. Indeed, using the fact that the product of Gaussians is a Gaussian and the moment matching conditions, we obtain the EP update rules for the variances $\bm{d}$ and the means $\bm{a}$:
\begin{equation}
    d_i=\left(\frac{1}{\langle h^2_i \rangle_{Q^{(i)}} - \langle h_i \rangle^2_{Q^{(i)}}}-\frac{1}{\Sigma^{(i)}_{ii}}\right)^{-1},
    \label{eq:EP_update1}
\end{equation}
\begin{equation}
    a_i=\langle h_i \rangle_{Q^{(i)}}+\frac{d_i}{\Sigma^{(i)}_{ii}}\left(\langle h_i \rangle_{Q^{(i)}}-\bar{h}^{(i)}_i\right),
    \label{eq:EP_update2}
\end{equation}
for all $i=1,\dots,N+M$.
Following \citep{braunstein2017,csep2020}, the cavity variances $\Sigma^{(i)}_{ii}$ and means $\bar{h}_i^{(i)}$ appearing in Eq. \eqref{eq:EP_update1} and \eqref{eq:EP_update2} can be computed in terms of the variances $\Sigma_{ii}$ and means $\bar{h}_i$ using a low rank update rule:
\begin{equation}
\Sigma^{(i)}_{ii}=\frac{\Sigma_{ii}}{1-\frac{1}{d_i}\Sigma_{ii}},
\label{eq:low-rank-update-sigma}
\end{equation}
\begin{equation}
\bar{h}_i^{(i)}=\frac{\bar{h}_i-\Sigma_{ii}\frac{a_i}{d_i}}{1-\frac{\Sigma_{ii}}{d_i}},
\label{eq:low-rank-update-mu}
\end{equation}
which allows to perform only one matrix inversion per iteration.

EP repeatedly estimates the vectors $\bm{a}$ and $\bm{d}$ until a fixed point is eventually reached. From a practical point of view, the algorithm returns the means and the variances of the marginal tilted distributions as soon as the convergence criterion:
\begin{equation}
    \varepsilon_t:=\max_i{\left\{\left|\langle h_i\rangle_{Q_t^{(i)}} - \langle h_i\rangle_{Q_{t-1}^{(i)}} \right| + \left|\langle h^2_i\rangle_{Q_t^{(i)}} - \langle h^2_i\rangle_{Q_{t-1}^{(i)}} \right|\right\}}<\varepsilon_{\text{stop}}
\end{equation}
 is fulfilled, where $t$ denotes the current iteration and $\varepsilon_{\text{stop}}$ is a convergence threshold. In particular, the posterior mean value of weights learned by the student perceptron are estimated as given by $\langle w_i \rangle_{Q^{(i)}}$, with a standard deviation equal to $\sqrt{\langle w^2_i \rangle_{Q^{(i)}}-\langle w_i \rangle^2_{Q^{(i)}}}$.

The zero temperature formulation of EP presented in this section is computationally advantageous compared to the finite temperature one presented in Appendix \ref{appendix: finite T EP}, as its complexity is dominated by the computation of the $N^2$ scalar products between vectors of length $M$ that appear in the second term of the right-hand side of Eq. \eqref{eq:cov_Q} and by the inversion of the $N\times N$ matrix given by the same equation, resulting in a cost $O(MN^2+N^3)$, rather than $O((N+M)\times(N+M))$. In general, in order to reduce the computational burden related to the inversion of the covariance matrix \eqref{eq:cov_Q}, we perform a Cholesky decomposition before inverting. For more details about the finite temperature formulation of EP, we refer to Appendix \ref{appendix: finite T EP}. 

\subsection{Moments of the tilted distributions}
\subsubsection{Moments of the spike-and-slab prior}

In this section, we shall compute the first and second moments of the leave-one-out distributions when the prior is of the spike-and-slab type. We recall the expression of the spike-and-slab prior already introduced in Eq. \eqref{eq:spike-and-slab} for the sake of convenience:
\begin{equation}
\Gamma(h_{k})=(1-\rho)\delta(h_{k})+\rho\sqrt{\frac{\lambda}{2\pi}}e^{-\frac{1}{2}\lambda h_{k}^{2}},\quad k=1,\dots,N.
\end{equation}
The marginal tilted distribution of each weight of the student perceptron reads:
\begin{equation}
Q^{(k)}(h_k)=\frac{1}{Z_{Q^{(k)}}}\tilde{Q}_{k}(h_k)\Gamma(h_{k}),
\label{eq:tilted_gamma}
\end{equation}
where we have introduced the marginalized cavity Gaussian distribution $\tilde{Q}_{k}$:
\begin{equation}
    \tilde{Q}_{k}(h_{k};\mu_{k},\Sigma_{k})=\frac{1}{\sqrt{2\pi\Sigma_{k}}}e^{-\frac{(h_{k}-\mu_{k})^{2}}{2\Sigma_{k}}}.
\label{eq:cavity-gaussian}
\end{equation}
From Eq. \eqref{eq:tilted_gamma}, computing the partition function of the tilted distribution $Q^{(k)}$ yields:
\begin{equation}
    Z_{Q^{(k)}}=(1-\rho)\frac{1}{\sqrt{2\pi\Sigma_{k}}}e^{-\frac{\mu_k^{2}}{2\Sigma_{k}}}+\frac{\rho}{\sqrt{2\pi}}\sqrt{\frac{\lambda}{1+\lambda\Sigma_{k}}}e^{-\frac{1}{2}\frac{\lambda\mu_{k}^{2}}{1+\lambda\Sigma_{k}}}.
\end{equation}
Finally, the first moment and the second moment of the same distribution are given by:
\begin{equation}
\langle h_{k}\rangle_{Q^{(k)}}=\frac{1}{Z_{Q^{(k)}}}\frac{\rho}{\sqrt{2\pi}}e^{-\frac{1}{2}\frac{\lambda\mu_k^{2}}{1+\lambda\Sigma_{k}}}\sqrt{\frac{\lambda}{1+\lambda\Sigma_{k}}}\frac{\mu_{k}}{1+\lambda\Sigma_{k}},
\end{equation}
and by:
\begin{equation}
\langle h_{k}^{2}\rangle_{Q^{(k)}}=\frac{1}{Z_{Q^{(k)}}}\frac{\rho}{\sqrt{2\pi}}e^{-\frac{1}{2}\frac{\lambda\mu_{k}^{2}}{1+\lambda\Sigma_{k}}}\sqrt{\frac{\lambda}{1+\lambda\Sigma_{k}}}\left(\frac{\Sigma_{k}+\lambda\Sigma_{k}^{2}+\mu_{k}^{2}}{(1+\lambda\Sigma_{k})^{2}}\right),
\end{equation}
respectively.

\subsubsection{Moments of the theta pseudoprior}
We now repeat the same reasoning for the case of the theta pseudoprior, which was defined as:
\begin{equation}
    \Lambda(h_k)=\Theta(h_k),\quad k=N+1,\dots,N+M.
\end{equation}
The associated tilted distribution of the $k$-th variable is given by:
\begin{equation}
Q^{(k)}(h_k)=\frac{1}{Z_{Q^{(k)}}}\tilde{Q}_{k}(h_k)\Lambda(h_{k}), \quad k=N+1,\dots,N+M,
\label{eq:tilted_lambda}
\end{equation}
where the expression for $\tilde{Q}_{k}$ is the same as in Eq. \eqref{eq:cavity-gaussian}. The normalization of \eqref{eq:tilted_lambda} is the partition function of the tilted distribution and reads:
\begin{equation}
    Z_{Q^{(k)}}=\frac{1}{2}\left[1+\text{erf}\left(\frac{\mu_k}{\sqrt{2\Sigma_k}}\right)\right],
\end{equation}
where $erf$ denotes the error function, defined as:
\begin{equation}
    \erf(x)=\frac{2}{\sqrt{\pi}}\int_0^x e^{-z^2}dz.
\end{equation}
Computing the first moment of the marginal tilted distribution leads to the expression:
\begin{equation}
    \langle h_k \rangle = \mu_k + \sqrt{\frac{\Sigma_k}{2\pi}}\frac{e^-\frac{\mu^2_k}{2\Sigma_k}}{\Phi\left(\frac{\mu_k}{\sqrt{\Sigma_k}}\right)}=\mu_k\left(1+\frac{R(\alpha_k)}{\alpha_k}\right),
\end{equation}
where $\Phi(x)=\frac{1}{2}\left[1+\erf\left(\frac{x}{\sqrt{2}}\right)\right]$ is the cumulative density function (CDF) of the standard normal distribution, $R(x)=\frac{1}{\sqrt{2\pi}}\frac{e^{-{x^2/2}}}{\Phi(x)}$ and $\alpha_k=\frac{\mu_k}{\sqrt{\Sigma_k}}$. Finally, concerning the second moment of the marginal tilted distribution, one obtains:
\begin{equation}
    \langle h^2_k \rangle = \mu^2_k+\Sigma_k+\mu_k\sqrt{\Sigma_k}R\left(\alpha_k\right),
\end{equation}
implying that the variance of $h_k$ w.r.t. the $k$-th marginal tilted distribution can be expressed in a compact way by:
\begin{equation}
    \text{Var}(h_k)=\Sigma_k(1-\alpha_k R(\alpha_k)-R^2(\alpha_k)).
\end{equation}

\subsubsection{Moments of the theta mixture pseudoprior}

When the pseudoprior $\Lambda$ is of the theta mixture type:
\begin{equation}
    \tilde{\Lambda}(h_k)=\eta\Theta(h_k)+(1-\eta)\Theta(-h_k), \quad 0\leq \eta\leq 1, \quad k=N+1,\dots,N+M,
    \label{eq:ThetaMixturePrior}
\end{equation}
we have for the partition function $Z_{Q^{(k)}}$ of the tilted distributions \eqref{eq:tilted_lambda}:
\begin{equation}
    Z_{Q^{(k)}}=\eta\left[\frac{1}{2}\erfc\left(-\frac{\mu_k}{\sqrt{2\Sigma_k}}\right)\right]+(1-\eta)\left[\frac{1}{2}\erfc\left(\frac{\mu_k}{\sqrt{2\Sigma_k}}\right)\right]=\sqrt{\frac{\pi\Sigma_{k}}{2}}\left[\frac{1}{2}+\left(\eta-\frac{1}{2}\right)\erf\left(\frac{\mu_k}{\sqrt{2\Sigma_k}}\right)\right].
\end{equation}
For the first moment, one obtains:
\begin{equation}
    \begin{split}
        \langle h_k \rangle_{Q^{(k)}} &= \frac{1}{Z_{Q^{(k)}}} \left\{\frac{\eta}{\sqrt{2\pi\Sigma_k}}\left[\Sigma_k e^{-\frac{\mu^2_k}{2\Sigma_k}}+\mu_k\sqrt{\frac{\pi\Sigma_k}{2}}\erfc\left(-\frac{\mu_k}{\sqrt{2\Sigma_k}}\right)\right] + \right.\\
        &+\left. \frac{1-\eta}{\sqrt{2\pi\Sigma_k}}\left[-e^{-\frac{\mu^2_k}{2\Sigma_k}}+\mu_k\sqrt{\frac{\pi\Sigma_k}{2}}\erfc\left(\frac{\mu_k}{\sqrt{2\Sigma_k}}\right)\right]\right\}=\\
        &= \mu_k+\sqrt{\frac{2\Sigma_k}{\pi}}\frac{(2\eta-1)e^{-\frac{\mu_k^2}{2\Sigma_k}}}{\eta\cdot \erfc\left(-\frac{\mu_k}{\sqrt{2\Sigma_k}}\right)+(1-\eta)\erfc\left(\frac{\mu_k}{\sqrt{2\Sigma_k}}\right)},
    \end{split}
\end{equation}
and the second moment w.r.t. the marginal tilted distribution \eqref{eq:tilted_lambda} reads:
\begin{equation}
\begin{split}
    \langle h^2_k \rangle_{Q^{(k)}} &= \frac{1}{Z_{Q^{(k)}}}\left\{\frac{\eta}{\sqrt{2\pi\Sigma_k}}\left[\mu_k\Sigma_k e^{-\frac{\mu_k^2}{2\Sigma_k}}+\sqrt{\frac{\pi\Sigma_k}{2}}\left(\mu_k^2+\Sigma_k\right)\erfc\left(-\frac{\mu_k}{\sqrt{2\Sigma_k}}\right)\right]+\right. \\
    &+\left.\frac{1-\eta}{\sqrt{2\pi\Sigma_k}}\left[-\mu_k\Sigma_k e^{-\frac{\mu_k^2}{2\Sigma_k}}+\sqrt{\frac{\pi\Sigma_k}{2}}\left(\mu_k^2+\Sigma_k\right)\erfc\left(\frac{\mu_k}{\sqrt{2\Sigma_k}}\right)\right]\right\}=\\
    &=\mu_k^2+\Sigma_k+\mu_k\sqrt{\frac{2\Sigma_k}{\pi}}\frac{(2\eta-1) e^{-\frac{\mu_k^2}{2\Sigma_k}}}{\eta\cdot\erfc\left(-\frac{\mu_k}{\sqrt{2\Sigma_k}}\right)+(1-\eta)\erfc\left(\frac{\mu_k}{\sqrt{2\Sigma_k}}\right)}.
    \end{split}
\end{equation}

\section{Results}
\label{sec:results}
\subsection{Sparse perceptron learning from noiseless examples}
In this section, we will present some results obtained from numerical
simulations in the presence of noiseless examples, both in the case
where patterns are independent and identically distributed (i.i.d.)
and in a simple case of correlated patterns. For the sake of
simplicity, in all the situations described in the following, we have
chosen a Bayes-optimal setting, where the prior information provided
by the spike-and-slab prior mirrors the actual distribution of the
weights to be retrieved.

First, we performed numerical experiments with i.i.d. patterns drawn
from a Gaussian distribution having zero mean and unit variance. As a
performance measure, we consider the mean squared error between the
normalized weights of the student perceptron at the end of the
learning process and those of the teacher perceptron:
\begin{equation}
\mathrm{MSE}(\boldsymbol{\tilde{w}},\boldsymbol{\tilde{B}})=\frac{1}{N}\sum_{k=1}^{N}{(\tilde{w}_k-\tilde{B}_k)^2},
\label{eq:MSE}
\end{equation}
where $\boldsymbol{\tilde{w}}=\boldsymbol{w}/\Vert \boldsymbol{w}
\Vert$ are the rescaled weights of the student and
$\boldsymbol{\tilde{B}}=\boldsymbol{B}/\Vert \boldsymbol{B} \Vert$
denote those of the teacher.  In our results, this metric is expressed
in decibel (dB) and used to compare expectation propagation to the
1-bit Approximate Message Passing (1bitAMP) algorithm introduced in
\cite{onebitAMP} and to the grVAMP algorithm proposed within the
unified Bayesian framework of general linear models published in
reference \cite{grVAMP}, which the authors show to yield equivalent
results to the VAMP algorithm for the generalized linear model
described in \cite{schniter2016vector}. We refer to
Fig. \ref{fig:1bitAMP-FG} and to Fig. \ref{fig:grVAMP-FG} for the
factor graphs related to the 1bitAMP and to the grVAMP approximations,
respectively. The computational cost of 1bitAMP is $O(N^2)$, while the
cost of grVAMP is $O(N^3)$, as it involves a one-time initial singular
value decomposition of $\bf X$. However, since this computation is not
needed again in the following part of the algorithm, its cost can be
neglected for small enough values of $N$. The remaining part of the
grVAMP scheme shares the same per iteration computational cost of
1bitAMP, as both are dominated by a matrix-vector product.

\begin{figure}
\includegraphics[width=0.6\textwidth]{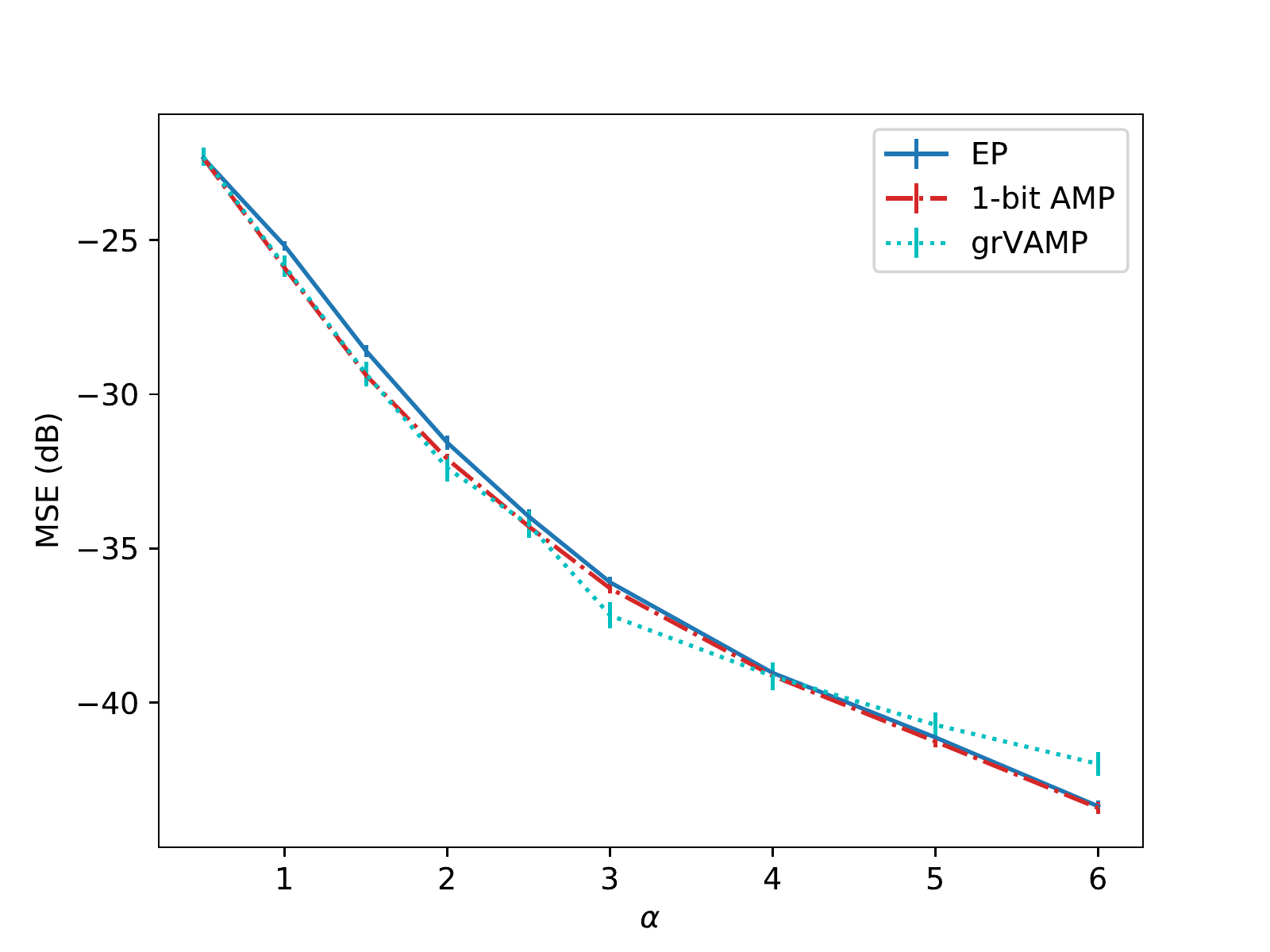}
\caption{$MSE$ resulting from sparse weight learning from i.i.d. patterns using EP,  1-bit AMP and grVAMP based estimation as a function of $\alpha$. The parameters considered for the perceptron are $N=128$ and $\rho=0.25$ and the number of instances is $N_{samples}=100$. All simulations converged and the MSE shown is averaged over all the considered instances. The error bars are estimated as $\sigma/\sqrt{N_{samples}}$, where $\sigma$ is the sample standard deviation of the $MSE$ computed over all the instances. }
\label{fig:iid_patterns}
\end{figure}

We considered the average of the MSE \eqref{eq:MSE} over
$N_{samples}=100$ simulations. The simulations correspond to sparse
perceptron learning of different instances of the weights of the
teacher perceptron, each from a different set of i.i.d. Gaussian
patterns fed to the student perceptron. We considered the case in
which the total number of weights is $N=128$ and their density level
is fixed to $\rho=0.25$. The Gaussian part of the spike and slab prior 
was set to a standard Gaussian distribution in 1bitAMP, EP and grVAMP. 
The EP convergence threshold was set to
$\epsilon_\text{stop}=10^{-4}$ and the value of the damping parameter of the EP
algorithm was set equal to 0.9995 (although good results can be
obtained using a lower damping too, e.g. 0.99). The results of the
simulations for different values of $\alpha$ are reported in Fig.
\ref{fig:iid_patterns} and show that EP, 1-bit AMP and grVAMP based
learning from i.i.d. Gaussian patterns have roughly the same
performance regardless of the specific value of $\alpha$.  The
convergence criterion was $10^{-4}$ in the 1-bit AMP simulations and
$10^{-8}$ in the grVAMP simulations. All the simulations performed 
using EP, 1-bit AMP and grVAMP converged within the thresholds we 
considered. The error bars in Fig. \ref{fig:iid_patterns} were estimated as
$\sigma/\sqrt{N_{samples}}$, where $\sigma$ denotes the sample standard 
deviation of the MSE.

\begin{figure}
      \begin{subfloat}[]{
      \includegraphics[width=0.47\textwidth]{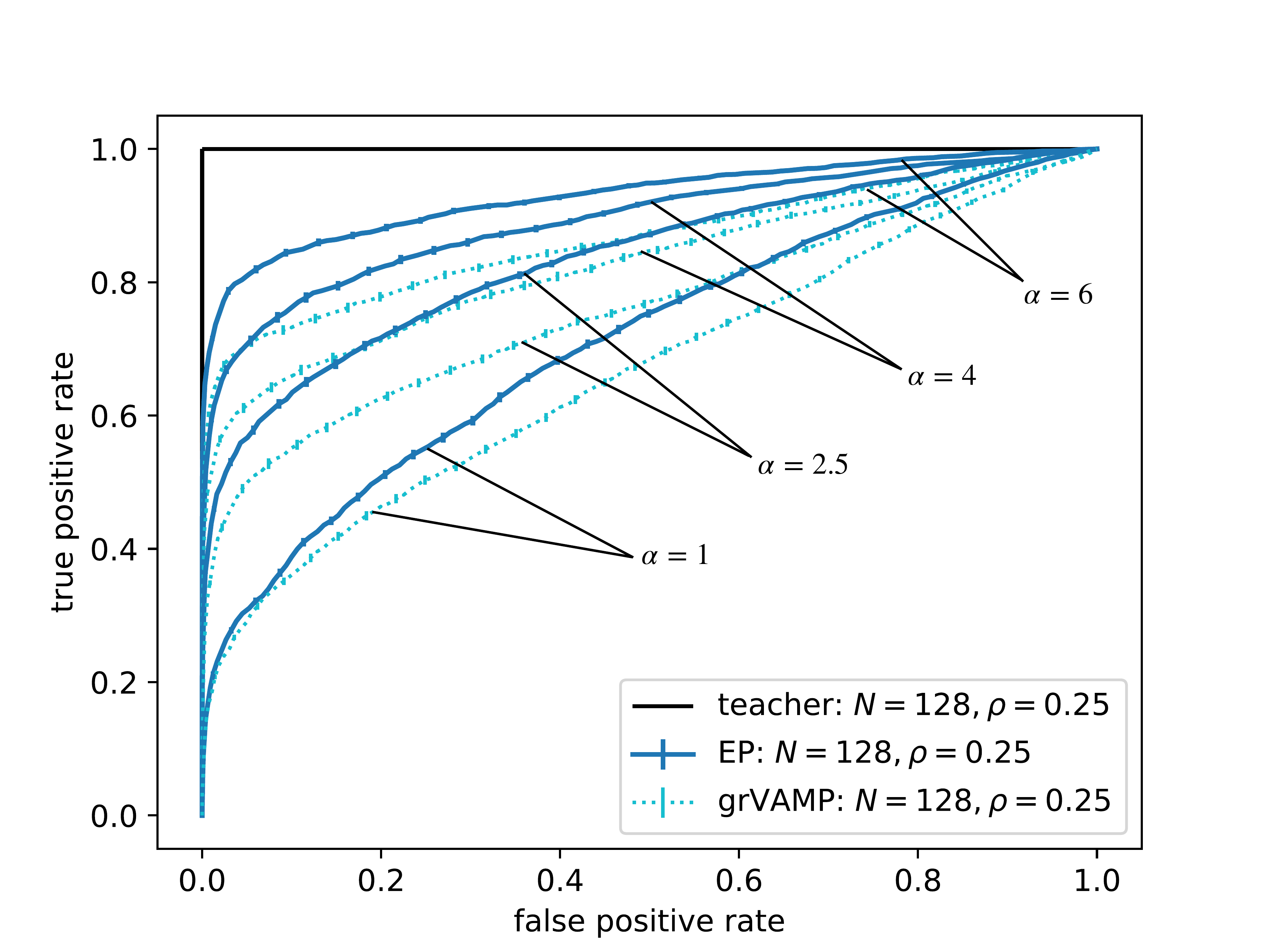}
      \label{fig:EP_vs_grVAMP_ROC_cor}}
      \end{subfloat}
      \hfill
      \begin{subfloat}[]{
      \includegraphics[width=0.47\textwidth]{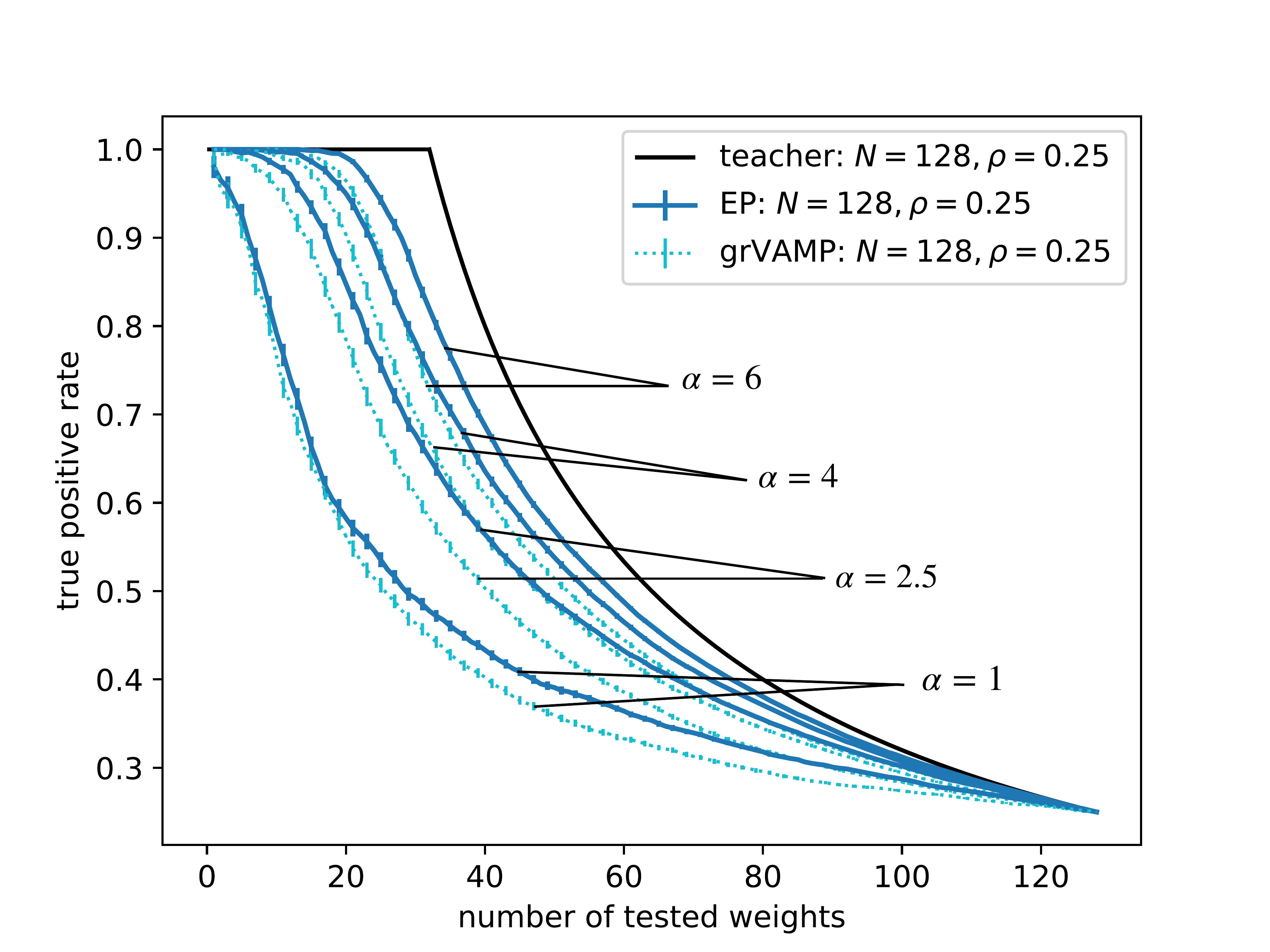}
      \label{fig:EP_vs_grVAMP_SensPlot_cor}}
      \end{subfloat}
      \\
      \begin{subfloat}[]{
      \includegraphics[width=0.6\textwidth]{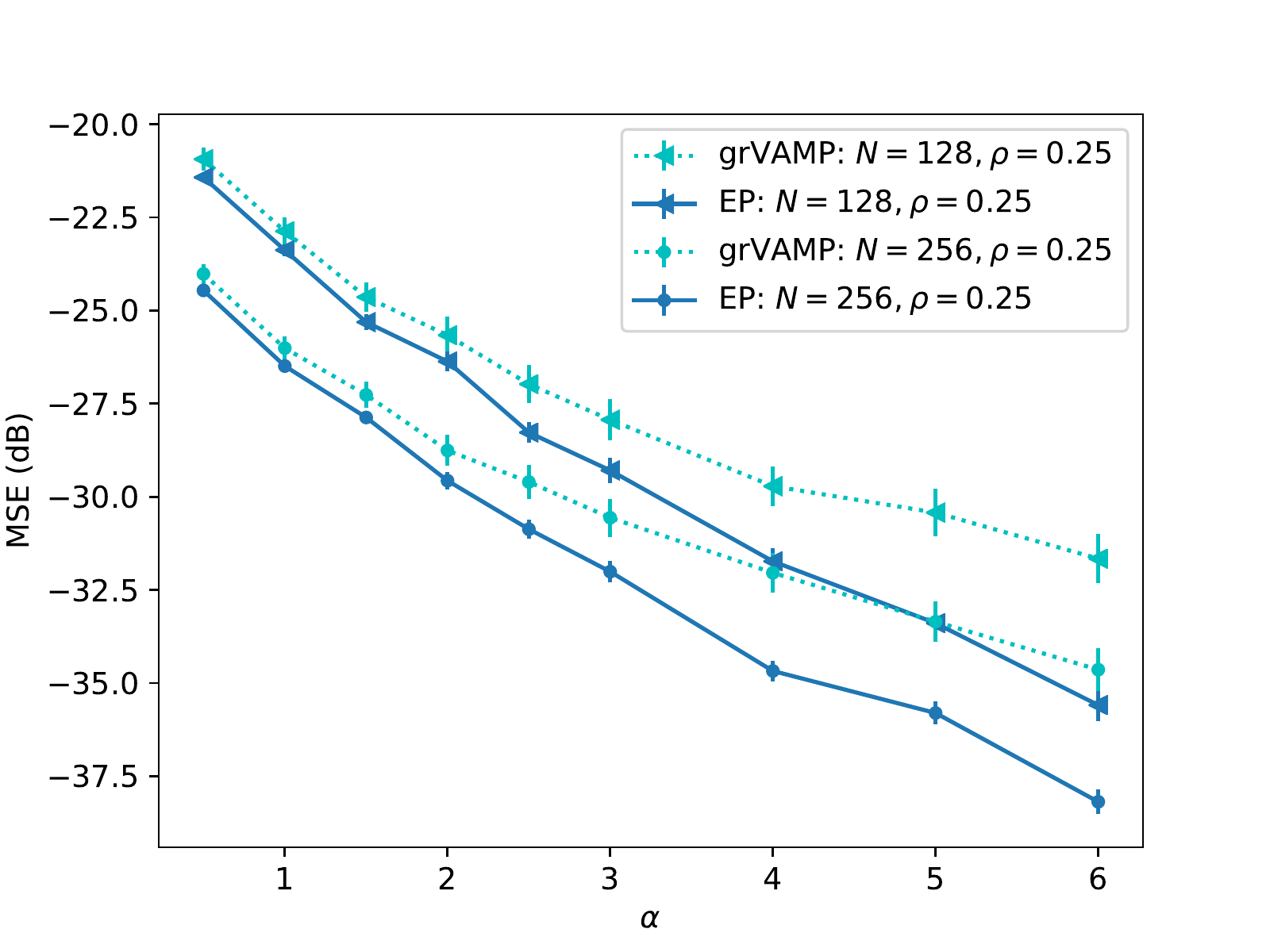}
      \label{fig:EP_vs_grVAMP_MSE_corr}}
      \end{subfloat}
      \caption{Sparse perceptron learning from correlated patterns
        sampled from a multivariate Gaussian distribution. 
        The values of the parameters are specified in each 
        panel and we set $u=1$. Comparison
        between grVAMP and EP based learning. (a) ROC curves. 
        (b) Sensitivity plots. For reference, in (a,b) the case of ideal variable selection
        by the teacher perceptron that provided the examples is also shown. 
        (c) Mean squared error in dB. In each plot, the mean values and  
        the standard deviations are computed over the set of all $N$
        instances for which convergence was achieved. The error bars are 
        estimated as $\sigma/\sqrt{N}$, where $\sigma$ is the sample
        standard deviation over the same set of instances.}
      \label{fig:EP_vs_grVAMP_cor}
\end{figure}

We also considered the problem of sparse perceptron learning from
correlated patterns drawn from a multivariate normal distribution, in
the simple case where the mean is $\boldsymbol m=\boldsymbol{0}$ and
the covariance matrix is constructed according to:
\begin{equation}
  \mathbf{S}=\mathbf{Y}^T \mathbf{Y}+\boldsymbol{\Delta},
  \label{eq: MVN covariance matrix}
\end{equation}
where $\mathbf{Y} \in \mathbb{R}^{u\times N}$ is an i.i.d. matrix with
entries drawn from a standard univariate Gaussian distribution and
$\boldsymbol{\Delta}$ is a diagonal matrix whose eigenvalues are given
by the absolute value of i.i.d. random entries drawn from the same
distribution. By construction, this matrix is symmetric and positive
definite and, therefore, is a proper covariance matrix. The diagonal
matrix $\boldsymbol{\Delta}$ is added in order to ensure that
$\mathbf{S}$ has full rank.  As an extreme case, we choose
$u=1$ for the matrix $\mathbf{Y}$.

\begin{table}
\centering
 \begin{subfloat}[$N=128$]{
	\begin{tabular}{|c|c|c|c|c|}
 \hline
 $\alpha$ & $f_{EP}$ & $\sigma_{f_{EP}}$ & $f_{grVAMP}$ & $\sigma_{grVAMP}$ \\
 \hline
 0.5 & 1 & 0  & 1 & 0\\
 \hline
 1.0 & 1 & 0 & 0.96 & 0.02 \\
 \hline
 1.5 & 1 & 0 & 0.89 & 0.03 \\
 \hline
 2.0 & 1 & 0 & 0.83 & 0.04 \\
 \hline
 2.5 & 1 & 0 & 0.87 & 0.03 \\
 \hline
 3.0 & 1 & 0 & 0.87 & 0.03 \\
 \hline
 4.0 & 1 & 0 & 0.85 & 0.04 \\
 \hline
 5.0 & 1 & 0 & 0.82 & 0.04 \\
 \hline
 6.0 & 1 & 0 & 0.80 & 0.04 \\
 \hline
 \end{tabular}
 \label{subtab:first}}
 \end{subfloat}
 \qquad\qquad
\begin{subfloat}[$N=256$]{
 	\begin{tabular}{|c|c|c|c|c|}
 \hline
 $\alpha$ & $f_{EP}$ & $\sigma_{f_{EP}}$ & $f_{grVAMP}$ & $\sigma_{grVAMP}$ \\
 \hline
 0.5 & 0.99 & 0.01  & 1 & 0\\
 \hline
 1.0 & 0.73 & 0.04 & 0.99 & 0.01 \\
 \hline
 1.5 & 0.92 & 0.03 & 0.94 & 0.02 \\
 \hline
 2.0 & 0.96 & 0.02 & 0.88 & 0.03 \\
 \hline
 2.5 & 0.88 & 0.03 & 0.83 & 0.04 \\
 \hline
 3.0 & 0.88 & 0.03 & 0.73 & 0.04 \\
 \hline
 4.0 & 0.92 & 0.03 & 0.80 & 0.04 \\
 \hline
 5.0 & 0.94 & 0.02 & 0.76 & 0.04 \\
 \hline
 6.0 & 0.96 & 0.02 & 0.70 & 0.05 \\
 \hline
 \end{tabular}
 \label{subtab:second}}
 \end{subfloat}
 \caption{Fraction of converged trials over a set of 100 different
   instances of the weights of the teacher perceptron and of the
   training set of examples. The patterns were sampled from the
   multivariate Gaussian distribution with covariance matrix
   \eqref{eq: MVN covariance matrix}. The number of variables is
   $N=128$ and the density of the weights of the teacher is
   $\rho=0.25$.}
 \label{tab: conv_EP_grVAMP_cor}
\end{table}

We find that even in this case the student perceptron is able to
estimate the weights of the teacher, although, under the same values
of the parameters $N$, $\rho$ and $\alpha$ of the model and under the 
same values of the EP parameters (i.e. damping, $\epsilon_\text{stop}$
and maximum number of iterations), 
the accuracy of the estimation is lower than the one
achieved by learning from i.i.d. Gaussian patterns, as one might
expect. Still, in the presence of the correlated patterns considered
here, expectation propagation based learning proves to be advantageous
as compared with other algorithms for 1-bit compressed sensing such as
1bitAMP, whose estimates of the means and variances of the weights
to be retrieved diverge. In addition, in the same situation, EP
outperforms grVAMP based learning, as shown in
fig. \ref{fig:EP_vs_grVAMP} for the set of parameters $N=128$ and
$\rho=0.25$.  The convergence thresholds of the grVAMP and of the EP
algorithms were set to the same values as in the case of learning from
i.i.d. Gaussian patterns and further lowering the value of the
threshold parameter of grVAMP did not result in a noticeable
improvement of the grVAMP results. In the case of EP, the damping 
factor was set to 0.999 and the maximum number of iterations for convergence 
was 50000. The fraction of converged trials is
shown in table \ref{tab: conv_EP_grVAMP_cor} for both algorithms in
the case where $N=128$ and $\rho=0.25$.  The EP led student perceptron
is more accurate at determining the nonzero weights than the grVAMP
led counterpart, as shown by the ROC curves in
Fig. \ref{fig:EP_vs_grVAMP_ROC_cor} and by the sensitivity plots of
Fig. \ref{fig:EP_vs_grVAMP_SensPlot_cor}. In order to construct
these curves, each weight of the teacher was assigned a score given by
its probability of being nonzero as estimated by EP and grVAMP. The
weights of the teacher were sorted in decreasing order according to these
probabilities, which are given by:
\begin{equation}
  P_{k}^{\neq0}=\left(1+\left(\frac{1}{\rho}-1\right)\sqrt{\frac{1+\lambda\Sigma_{k}}{\lambda\Sigma_{k}}}e^{-\frac{\mu_{k}^{2}}{2\Sigma_{k}(1+\lambda\Sigma_{k})}}\right)^{-1}.
  \label{P_nonzero}
\end{equation}
In the case of EP, $\mu_k$ and $\Sigma_k$, for $k=1,\dots,N$, are the
EP cavity means and variances, whereas, in the case of grVAMP,
$\boldsymbol{\mu}$ corresponds to the VAMP quantity $\mathbf{r}_{1k}$ and
$\Sigma_k=\gamma_{1k}^{-1}$, where $\gamma_{1k}$ is the quantity that
parameterizes the denoiser in VAMP \cite{VAMP} and we used the standard VAMP notation for  $\mathbf{r}_{1k}$ and $\gamma_{1k}$, for which the index $k$ refers to the number of the current iteration. In both cases, $\rho$ denotes the density
parameter of the spike-and-slab prior.  Interestingly,
the discrepancy between the accuracy of the two algorithms becomes
larger as the number of patterns increases and, as a consequence, the difference between the mean squared errors of the two
algorithms increases, as shown in Fig. \ref{fig:EP_vs_grVAMP_MSE_corr}, implying that the EP and the grVAMP approximations are very different in this Gaussian correlated pattern regime. This fact is confirmed by the heterogeneity of the variances $d_k,\quad k=1,\dots,N$ of the approximating univariate Gaussian factors $\phi_k$ when one considers the EP solution for instances where both EP and grVAMP converge. More precisely, it can be seen that the parameters $d_k$ span several orders of magnitude, contrary to VAMP where these variances are constrained to be equal.
In each plot in Fig. \ref{fig:EP_vs_grVAMP_cor}, we have shown the average of the 
quantities considered over the set of the $N_{conv}$ instances for 
which each algorithm achieved convergence. Accordingly, the error bars were estimated as
$\sigma/\sqrt{N_{conv}}$, where $\sigma$ denotes the standard deviation over 
the same set of instances.

A useful additional feature of the EP-based learning approach is the
possibility to learn iteratively the value of $\rho$ during the
estimation of the weights of the teacher, as, unlike EP, many algorithms for 1-bit
compressed sensing assume the density of the signal to be
given a priori. The estimation of the density parameter is achieved by
minimizing the EP free energy with respect to $\rho$ and yields good
results as long as the number of the patterns presented to the student
is large enough. We refer to the Appendix for details concerning the
EP free energy, its expression for the sparse perceptron learning
problem and free energy optimization based learning of the parameters
of the prior. We mention here that a similar expectation maximization based strategy can be implemented also in the case of 1bitAMP and grVAMP in order to estimate the density parameter $\rho$.

In order to show that our approach allows to estimate the dilution
level of the teacher perceptron, we performed a set of
$N_{samples}=100$ EP simulations on a system with $N=128$ and
$\rho=0.25$, where the density parameter $\rho_0$ of the
spike-and-slab prior assigned to each weight variable was randomly
initialized by sampling its value from a uniform distribution over the
interval $0.05\leq\rho_0\leq 0.95$ and where the learning rate was
chosen to be $\delta\rho=10^{-5}$. We show our results in table
\ref{tab: learning of rho}. For each value of $\alpha$, we show the
average value $\rho_L$ of the density estimate over all samples and
its associated statistical uncertainty, which was computed as
$\delta\rho_L=\sigma_{\rho_L}/\sqrt{N_{samples}}$, as for these values of
the parameters all simulations converged. We also show the
relative difference $\Delta\rho/\rho$ between the true value of the
density and the estimated one. Since $\Delta\rho\gg \delta\rho_L$, we
omit the statistical uncertainty associated with
$\Delta\rho$. Finally, we notice that, even when learning from
correlated patterns constructed as described above, the student is
able to estimate the density level of the weights of the teacher
perceptron quite accurately, provided that a sufficient number of
patterns is provided to the student perceptron. In table \ref{tab:
  learning of rho}, we give an example of this fact when the teacher
perceptron has $N=128$ weights and density $\rho=0.25$.

\begin{table}
\centering
\begin{tabular}{|c|c|c|c|c|}
\hline
$\alpha$ & i.i.d. patterns: $\rho_L\pm\delta\rho_L$ & i.i.d. patterns: $\Delta\rho/\rho$ & patterns from MVN: 		$\rho_L\pm\delta\rho_L$ & patterns from MVN: $\Delta\rho/\rho$ \\
\hline
2 & $0.191 \pm 0.003$ & $0.236$ & $0.161\pm 0.004$ & $0.341$ \\
\hline
3 & $0.220 \pm 0.002$ & $0.121$ & $0.196 \pm 0.004$ & $0.206$\\
\hline
4 & $0.234\pm0.002$ & $0.066$ & $0.207 \pm 0.003$ & $0.182$ \\
\hline
5 & $0.240 \pm 0.002$ & $0.042$ & $0.214 \pm 0.003$ & $0.144$\\
\hline
6 & $0.242 \pm 0.001$ & $0.031$ & $0.223 \pm 0.003$ & $0.115$\\
\hline
\end{tabular}
\caption{Learning of the density $\rho$ of the weights of the teacher 
for a perceptron with parameters $N=128$ and $\rho=0.25$. The average and the standard deviation of the learned value of $\rho$ at
  convergence over all the trials for which convergence was achieved
  during the training process are denoted by $\rho_L$ and $\delta\rho_L$, 
  respectively. In each trial, the initial condition $\rho_0$
  was drawn uniformly from the interval $0.05\leq\rho_0\leq 0.95$.}
\label{tab: learning of rho}
\end{table}

\subsection{Sparse perceptron learning from a noisy teacher}
\label{subsec: noisy perceptron}

We analyzed the performance of EP based sparse perceptron learning
when a small fraction of the examples is mislabeled. The student perceptron
is given the a priori information that a certain fraction of the
labels is wrongly assigned. As in the noiseless case, we consider a
Bayes-optimal setting and therefore we provide the student with such a priori knowledge using the theta mixture pseudoprior introduced in Eq. \eqref{eq:ThetaMixturePrior}.

In order to test and evaluate the performance of EP-guided learning in
this situation, we compared EP with grVAMP, in which we introduced a theta mixture measure as done in EP, and with the R1BCS algorithm for 1-bit
compressed sensing with sign-flip errors proposed by Li et
al. \cite{R1BCS}, which is based on an expectation maximization scheme
involving both the signal to be retrieved and the noise. In grVAMP, the slab part of the spike-and-slab prior was set to a standard Gaussian distribution. When using
R1BCS, we rescaled the pattern matrix so that each column had unit
norm and we included a convergence threshold
$\varepsilon_{R1BCS}=10^{-4}$. Thus, the R1BCS iterations stop when
the estimate $\boldsymbol{w}_{R1BCS}$ of the weights of the teacher is such
that
$\lVert\boldsymbol{w}_{R1BCS}-\boldsymbol{w}_{R1BCS}^{old}\rVert<\varepsilon_{R1BCS}$.
In each experiment, a given number $K_{label}=(1-\eta)M$ of labels were
flipped, where $\eta$ is the fraction of unchanged labels. In the EP simulations a damping factor equal to
0.99 and a convergence threshold $\epsilon_{\text{stop}}=10^{-6}$ were used. The parameter $\lambda$ of the spike-and-slab prior was set equal to $10^4$.

\begin{figure}
      \centering
      \begin{subfloat}[]{
      \includegraphics[width=0.46\textwidth]{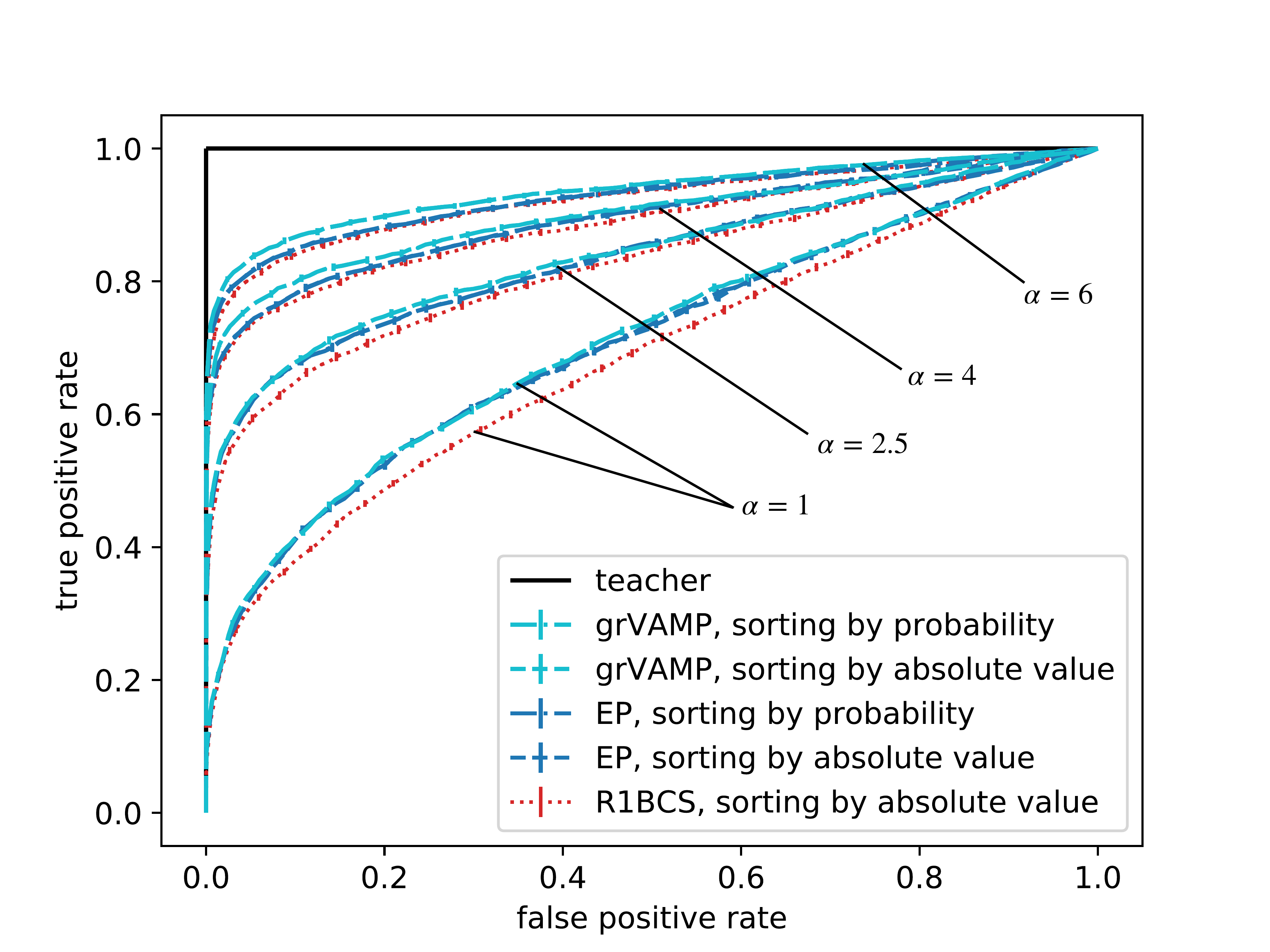}
      \label{fig:EP_vs_R1BCS_ROC_iid}}
      \end{subfloat}
      \hfill
      \begin{subfloat}[]{
      \includegraphics[width=0.46\textwidth]{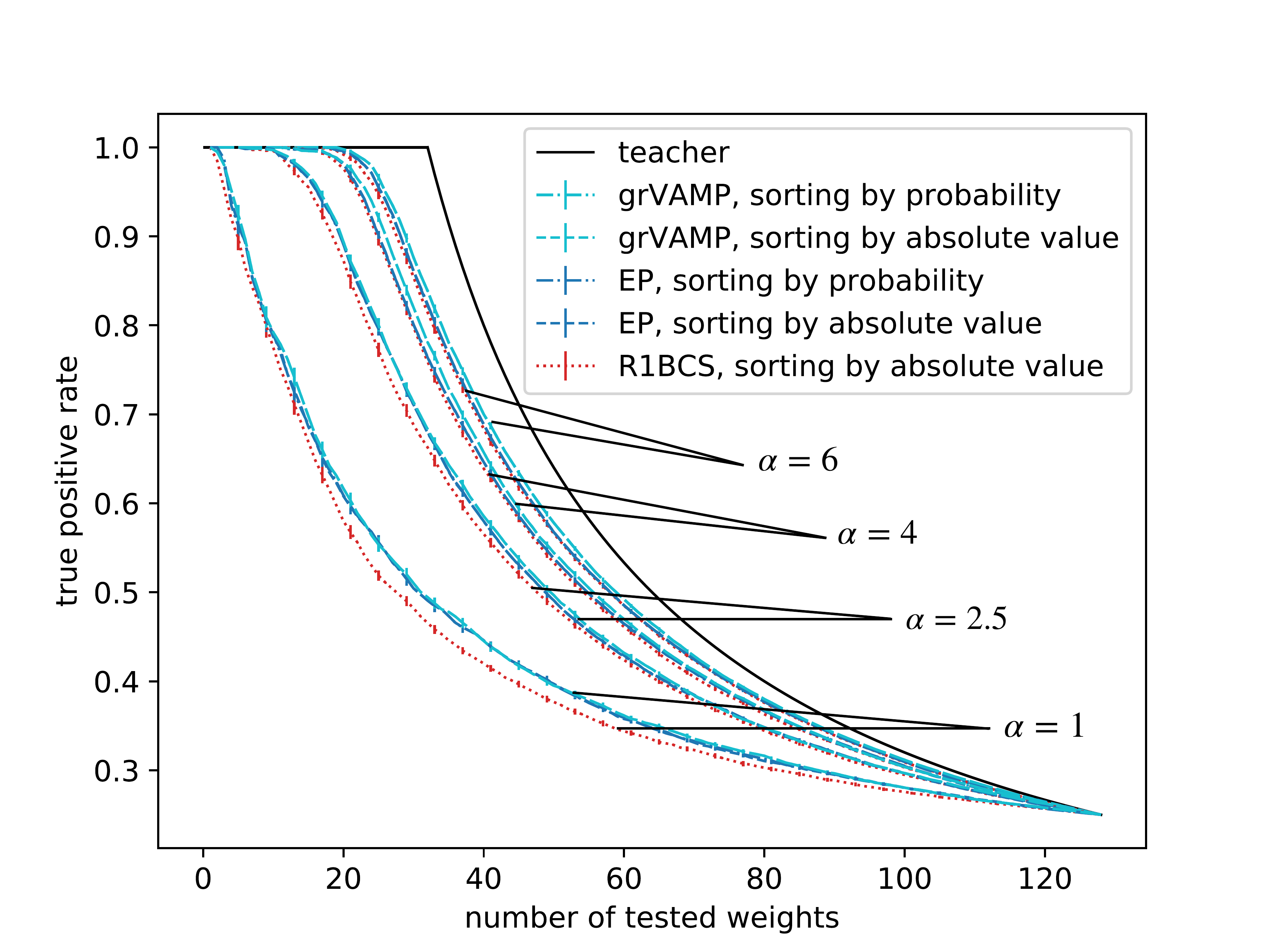}
      \label{fig:EP_vs_R1BCS_SP_iid}}
      \end{subfloat}
      \\
      \begin{subfloat}[]{
      \includegraphics[valign=c,width=0.46\textwidth]{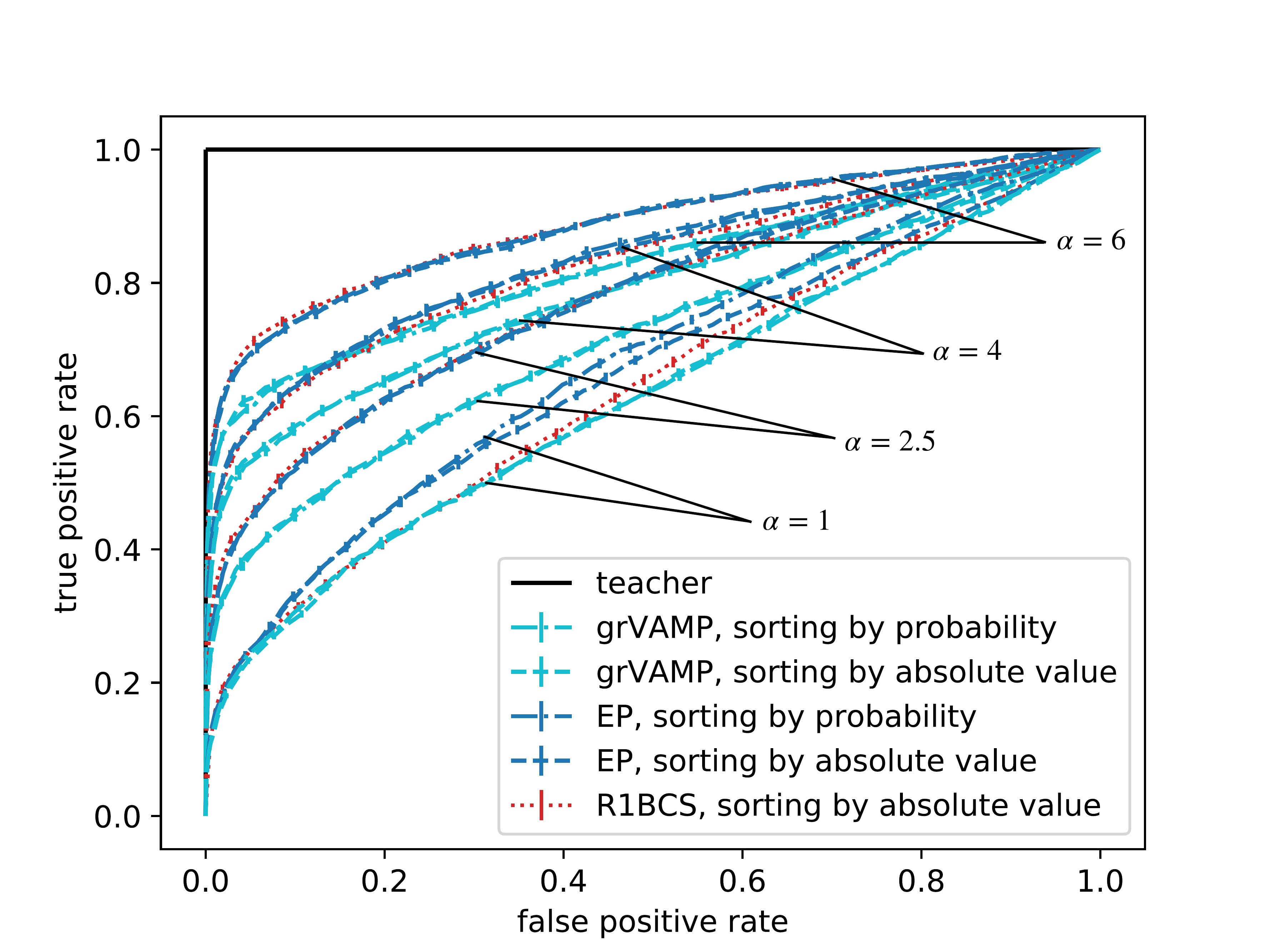}
      \label{fig:EP_vs_R1BCS_ROC_cor}}
      \end{subfloat}
      \hfill
      \begin{subfloat}[]{
      \includegraphics[valign=c,width=0.47\textwidth]{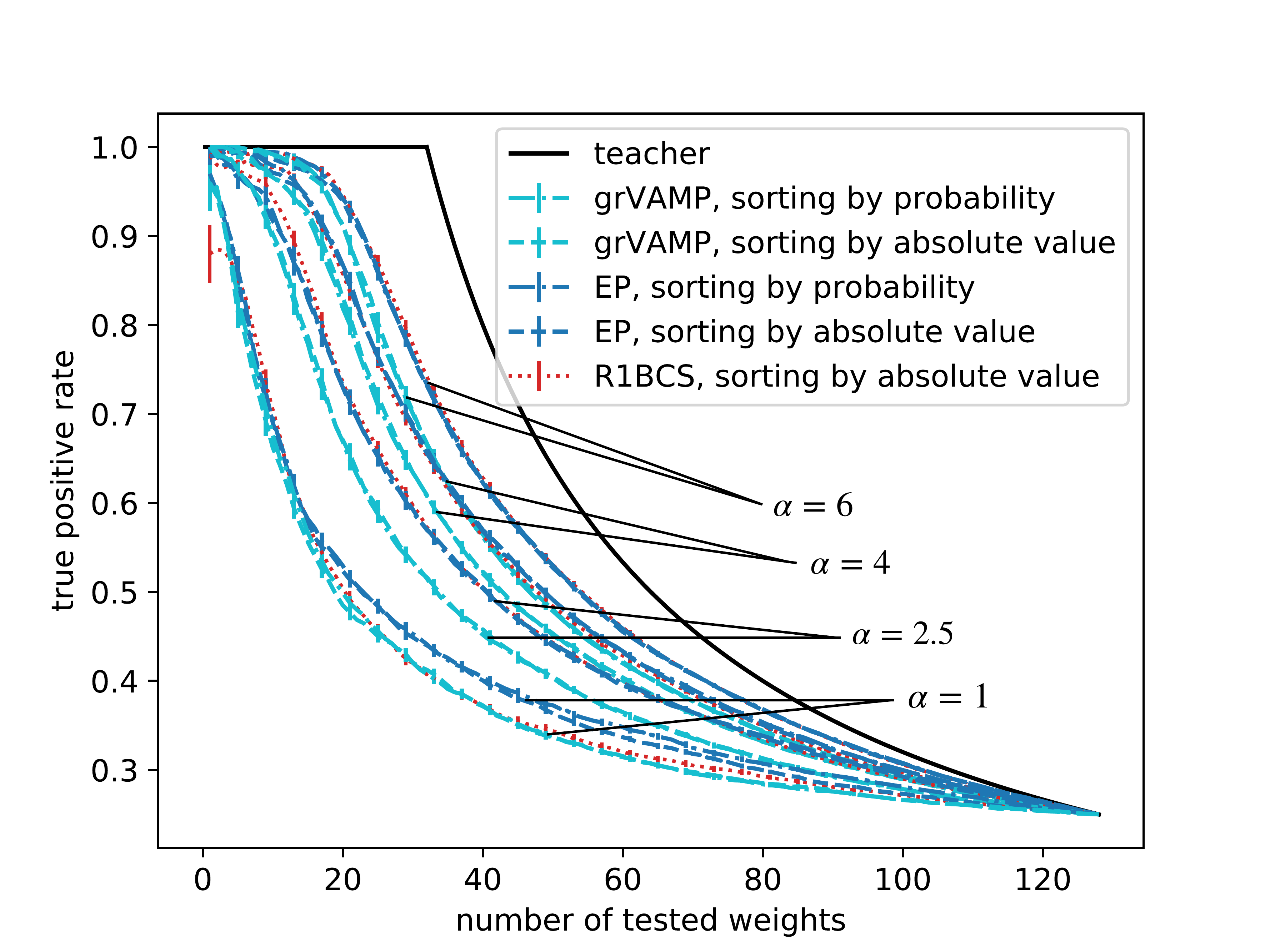}
      \label{fig:EP_vs_R1BCS_SP_cor}}
      \end{subfloat}
      \caption{Sparse perceptron learning from i.i.d. patterns sampled from a standard Gaussian distribution and from correlated patterns sampled from a multivariate Gaussian distribution, with $N=128$, $\rho=0.25$, $u=1$, $\eta=0.95$. A fraction $(1-\eta)$ of the labels are mislabeled. Comparison between R1BCS, grVAMP and EP in terms of their ROC curves (a,c) and of their sensitivity plots (b,d). For reference, the case of ideal variable selection by the teacher perceptron that provided the examples is shown. The plotted quantities are the mean values computed over the set of all $N$ instances and the error bars are estimated as $\sigma/\sqrt{N}$ where $\sigma$ denotes the sample standard deviation over all the instances considered.
     }
      \label{fig:EP_vs_R1BCS_iid}
\end{figure}

We first considered the case of i.i.d. Gaussian patterns drawn from a
Gaussian distribution with zero mean and unit variance and, subsequently, the case of correlated patterns drawn from the zero
mean multivariate Gaussian distribution with the covariance matrix expressed in 
\eqref{eq: MVN covariance matrix}, which we already introduced in
the noiseless case. For both kinds of patterns, we conducted numerical experiments on a set of 100 different instances. In the case of i.i.d patterns, convergence over all instances was achieved for all three algorithms within the convergence thresholds specified. The same was true in the case of correlated patterns for R1BCS and for EP, whereas grVAMP exhibited a failure rate up to $15\%$ in terms of convergence in the correlated case. 

\begin{figure}
	\centering
      \begin{subfloat}[]{
      \includegraphics[valign=c,width=0.46\textwidth]{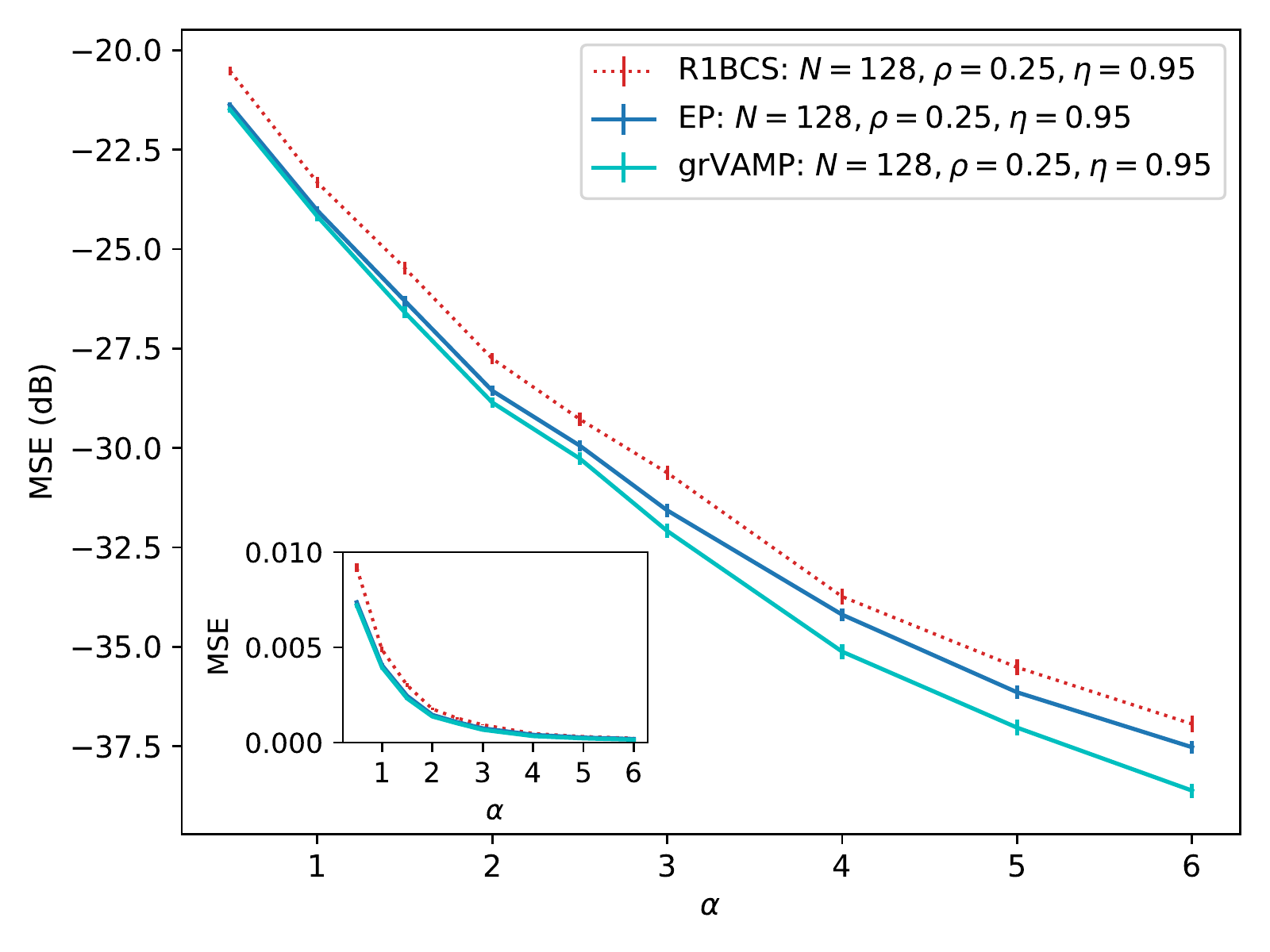}
      \label{fig:MSE_noise_iid_0.95}}
      \end{subfloat}
      \hfill
      \begin{subfloat}[]{
      \includegraphics[valign=c,width=0.46\textwidth]{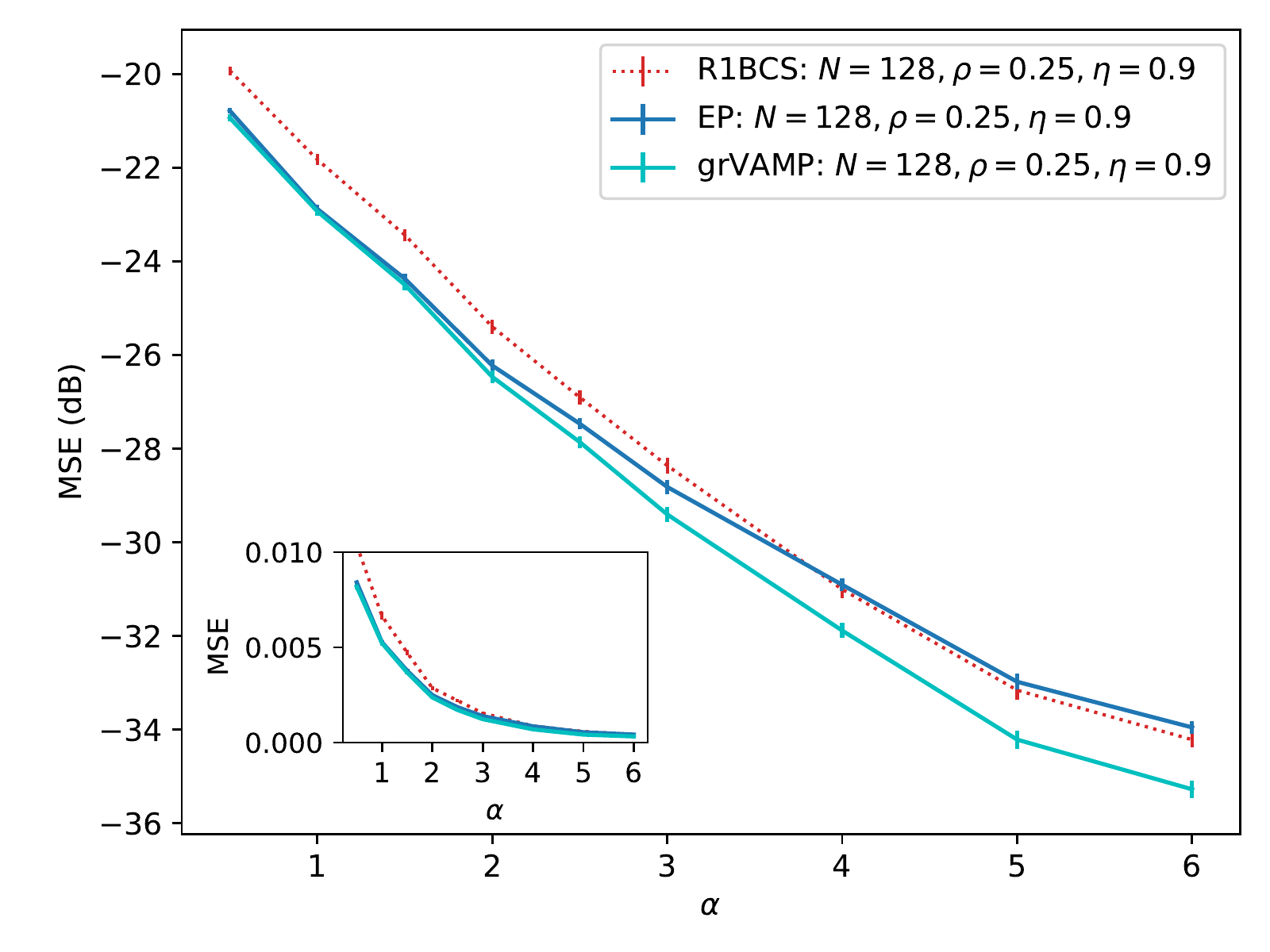}
      \label{fig:MSE_noise_iid_0.9}}
      \end{subfloat}
      \\
     \begin{subfloat}[]{
      \includegraphics[valign=c,width=0.46\textwidth]{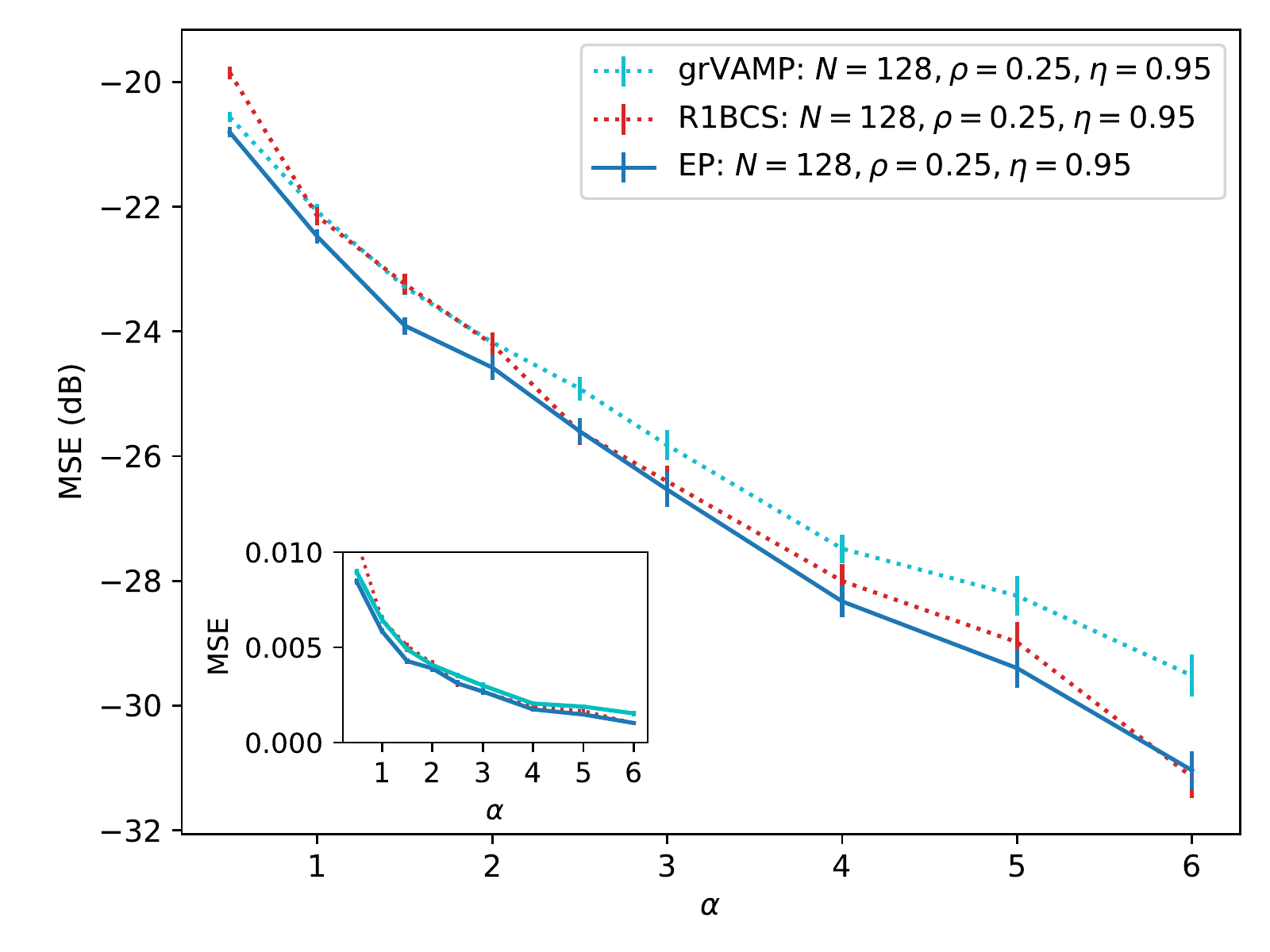}
      \label{fig:MSE_noise_cor_0.95}}
      \end{subfloat}
      \hfill
      \begin{subfloat}[]{
      \includegraphics[valign=c,width=0.46\textwidth]{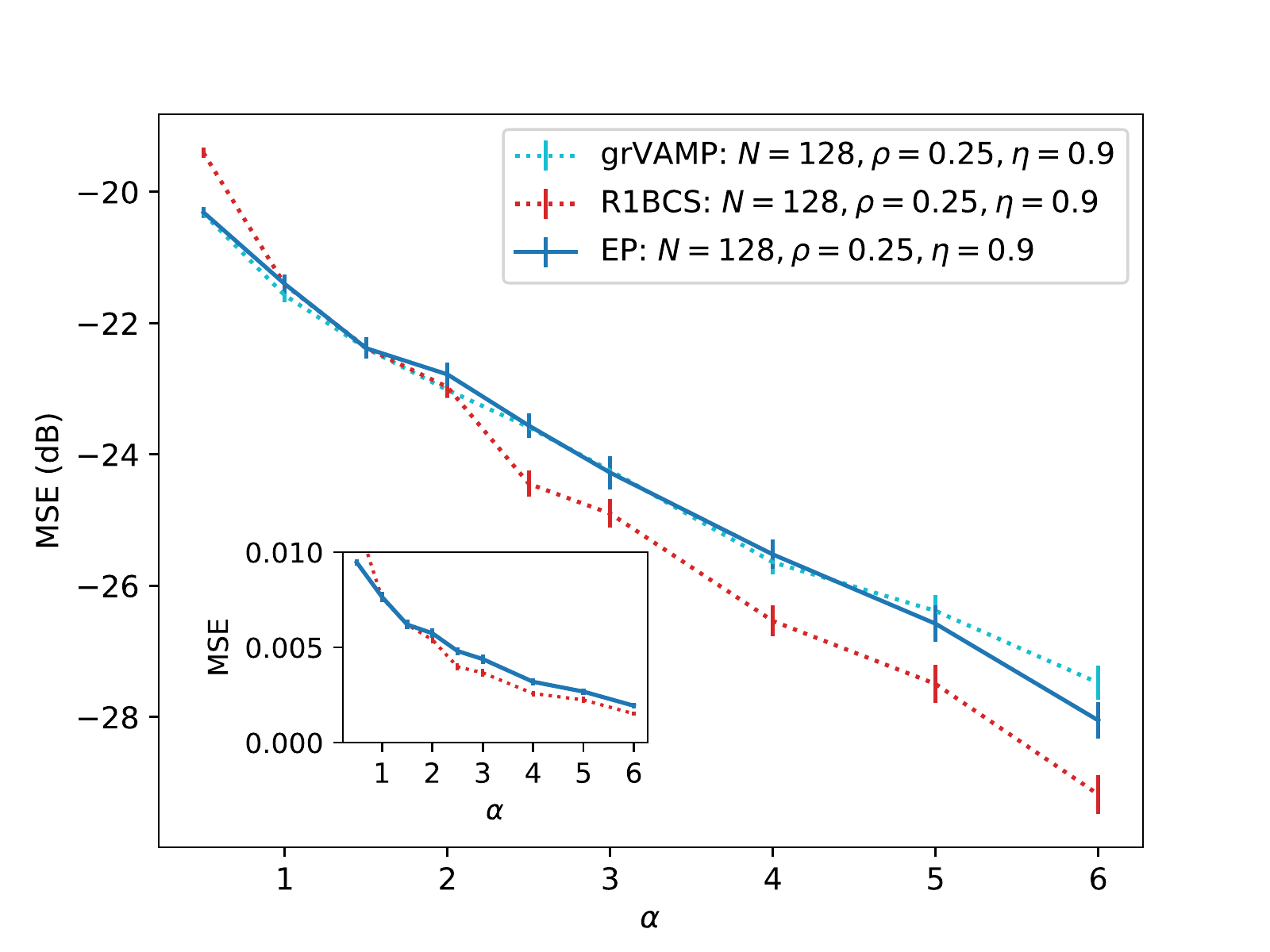}
      \label{fig:MSE_noise_cor_0.9}}
      \end{subfloat}
	\caption{Sparse perceptron learning from $(1-\eta)M$ mislabeled examples:  comparison between EP, grVAMP and R1BCS in terms of their mean squared errors in the case of (a) i.i.d. Gaussian patterns, $\eta=0.95$, (b) i.i.d. Gaussian patterns, $\eta=0.9$, (c) Gaussian correlated patterns, $\eta=0.95$ and (d) Gaussian correlated patterns, $\eta=0.9$. In all figures, $N=128$ and $\rho=0.25$. The mean squared errors plotted are averaged over the set of all $N$ instances and the error bars are estimated as $\sigma/\sqrt{N}$ where $\sigma$ denotes the sample standard deviation over all the instances considered.}
    \label{fig:MSE-noisy-case}
\end{figure}

In order to assess the variable selection
capabilities of EP when attempting to learn the
weights of the teacher from these kinds of patterns in the presence of
mislabeled examples, we computed the ROC curves and the sensitivity plots related to both the former (Fig. \ref{fig:EP_vs_R1BCS_ROC_iid} and \ref{fig:EP_vs_R1BCS_SP_iid}, respectively) and to the latter kind of patterns (Fig. \ref{fig:EP_vs_R1BCS_ROC_cor} and \ref{fig:EP_vs_R1BCS_SP_cor}, respectively). Such ROC curves and sensitivity plots are associated with the weights of the
student after the training phase was completed  and were obtained in the case where the number of weights was $N=128$, the density of the weights 
$\boldsymbol{B}$ of the teacher was $\rho=0.25$ and the fraction of uncorrupted labels was $\eta=0.95$.  The ordering
criterion for the weights $\boldsymbol{B}$ adopted in the ROC curves in
Fig. \ref{fig:EP_vs_R1BCS_ROC_iid} and \ref{fig:EP_vs_R1BCS_ROC_cor} and in the sensitivity plots in
Fig. \ref{fig:EP_vs_R1BCS_SP_iid} and \ref{fig:EP_vs_R1BCS_SP_cor} was based on the absolute value of
the weights $\boldsymbol{w}$ of the student. In the case of EP and grVAMP, we
also plotted the ROC and sensitivity curves according to the sorting
criterion based on the score expressed in Eq. \eqref{P_nonzero},
but the results did not exhibit noticeable differences with respect
to those obtained using the previous sorting criterion, as the curves in Fig. \ref{fig:EP_vs_R1BCS_iid} mostly overlap. 
The ROC curves and sensitivity plots associated with EP and grVAMP mostly exhibit similar values for the true positive ratio -- except for large $\alpha$, where the EP values are slightly smaller -- and outperform significatively the ones related to R1BCS in the case of i.i.d. patterns, especially at low values of $\alpha$. This is confirmed by the values of the areas under the curves, as shown in Table \ref{subtab:AUC_EP_vs_R1BCS_iid}, where the maximum relative discrepancy between the AUC is 0.008 when considering EP and grVAMP and 0.03 when considering EP and R1BCS. However, in the case of correlated patterns, the ROC curves and sensitivity plots related to EP are mostly comparable to those associated with R1BCS and yield values of the true positive rate that are systematically larger than the ones of grVAMP, implying that the variable selection properties of EP tend to be affected to a far lesser extent than those of grVAMP in this regime, as confirmed by the AUC reported in table \ref{subtab:AUC_EP_vs_R1BCS_cor} and analogously to what was observed in the noiseless case. In this case, the maximum relative discrepancy between the AUC of EP and of R1BCS is 0.05 and is attained at the lowest value of $\alpha$, namely $\alpha=0.5$, whereas, when considering EP and grVAMP, the relative discrepancy is largest at large values of $\alpha$, its maximum being 0.06 at $\alpha=6$. The mean squared error values associated with the normalized student and teacher are plotted in Fig. \ref{fig:MSE-noisy-case} for both $\eta=0.95$ and $\eta=0.9$ and their values  are shown both in dB (main plots) and using a linear scale (insets). In particular, the MSE values reported in Fig. \ref{fig:MSE_noise_iid_0.95} correspond to the ROC curves in Fig. \ref{fig:EP_vs_R1BCS_ROC_iid}. While the MSE of EP is definitely larger than that of grVAMP at large $\alpha$ in the presence of i.i.d. patterns (Fig. \ref{fig:MSE_noise_iid_0.95} and Fig. \ref{fig:MSE_noise_iid_0.9}), it should be noted that the differences of the values involved are very small and are not noticeable on a linear scale. Since the EP approximation is richer than the VAMP one, which, in turn, is due to the fact that EP reduces to VAMP when the approximate univariate prior factors are constrained to have the same variance, and because of the very small errors and differences observed, we believe that such observed discrepancy between grVAMP and EP in the i.i.d. pattern regime at large $\alpha$ is due to numerical effects rather than to intrinsic limitations of the EP scheme. Instead, this is not the case for the differences displayed by the MSE related to EP and grVAMP in Fig. \ref{fig:MSE_noise_cor_0.95} for $\eta=0.95$, which are clearly noticeable on a linear scale as well. The discrepancy involved reflects that observed at the level of the ROC curves shown in Fig. \ref{fig:EP_vs_R1BCS_ROC_cor}. Finally, in the Gaussian correlated scenario, as soon as the noise level affecting the labels becomes large enough, we see that EP and grVAMP yield similar results at all values of $\alpha$, as shown in Fig. \ref{fig:MSE_noise_cor_0.9}. 

Similarly to R1BCS, EP was able to correctly estimate the noise level affecting the
labels. This was achieved by sampling the initial condition $\eta_0$
for the parameter $\eta$ of the theta mixture pseudoprior uniformly from the
interval $0.5<\eta<1$ and performing one step of gradient descent on
the EP free energy at each EP iteration as described in Appendix
\ref{app:prior_param_learning}. In addition, analogously to the noiseless case, it is possible to learn the $\rho$ parameter and we verified that the two parameters can be learned simultaneously. We show the estimated values of the density of the weights and of the consistency level $\eta$ of the labels in the case of i.i.d. patterns in table \ref{subtab:learned_eta_iid} and in the case of Gaussian correlated patterns in table \ref{subtab:learned_eta_cor}. We notice that, analogously to EP, an expectation maximization scheme can be implemented in the case of grVAMP in order to iteratively learn the density parameter $\rho$ of the spike-and-slab prior and the $\eta$ parameter of the theta mixture measure.

\begin{table}[tb]
  \centering
  \begin{subfloat}[AUC (i.i.d. patterns) \label{subtab:AUC_EP_vs_R1BCS_iid}]{
  		\footnotesize
  		\centering
		\begin{tabular}{|c|c|c|c|}
		\hline
		$\alpha$ & $AUC_{EP}$ & $AUC_{grVAMP}$ & $AUC_{R1BCS}$\\
		\hline
$ 0.5 $ &$ 0.621 \pm 0.005 $ & $ 0.627 \pm 0.005 $ & $ 0.595 \pm 0.005 $ \\
\hline
$ 1.0 $ &$ 0.706 \pm 0.005 $ & $ 0.710 \pm 0.005 $ & $ 0.682 \pm 0.004 $ \\
\hline
$ 1.5 $ &$ 0.770 \pm 0.005 $ & $ 0.777 \pm 0.005 $ & $ 0.746 \pm 0.005 $ \\
\hline
$ 2.0 $ &$ 0.806 \pm 0.005 $ & $ 0.809 \pm 0.005 $ & $ 0.792 \pm 0.004 $ \\
\hline
$ 2.5 $ &$ 0.835 \pm 0.004 $ & $ 0.840 \pm 0.004 $ & $ 0.824 \pm 0.005 $ \\
\hline
$ 3.0 $ &$ 0.860 \pm 0.004 $ & $ 0.865 \pm 0.005 $ & $ 0.854 \pm 0.004 $ \\
\hline
$ 4.0 $ &$ 0.893 \pm 0.004 $ & $ 0.899 \pm 0.004 $ & $ 0.887 \pm 0.004 $ \\
\hline
$ 5.0 $ &$ 0.913 \pm 0.004 $ & $ 0.920 \pm 0.004 $ & $ 0.91 \pm 0.004 $ \\
\hline
$ 6.0 $ &$ 0.927 \pm 0.003 $ & $ 0.936 \pm 0.003 $ & $ 0.923 \pm 0.003 $ \\
\hline
		\end{tabular}}
  \end{subfloat}
  \qquad
  \begin{subfloat}[AUC (patterns from MVN) \label{subtab:AUC_EP_vs_R1BCS_cor}]{
  		\footnotesize
  		\centering
		\begin{tabular}{|c|c|c|c|}
		\hline
		$\alpha$ & $AUC_{EP}$ & $AUC_{grVAMP}$ & $AUC_{R1BCS}$\\
		\hline
		$ 0.5 $ &$ 0.588 \pm 0.006 $ & $ 0.579 \pm 0.006 $ & $ 0.559 \pm 0.005 $ \\
\hline
$ 1.0 $ &$ 0.661 \pm 0.005 $ & $ 0.628 \pm 0.006 $ & $ 0.641 \pm 0.006 $ \\
\hline
$ 1.5 $ &$ 0.727 \pm 0.006 $ & $ 0.685 \pm 0.006 $ & $ 0.694 \pm 0.006 $ \\
\hline
$ 2.0 $ &$ 0.732 \pm 0.007 $ & $ 0.696 \pm 0.006 $ & $ 0.734 \pm 0.006 $ \\
\hline
$ 2.5 $ &$ 0.775 \pm 0.007 $ & $ 0.719 \pm 0.006 $ & $ 0.775 \pm 0.006 $ \\
\hline
$ 3.0 $ &$ 0.788 \pm 0.007 $ & $ 0.742 \pm 0.007 $ & $ 0.793 \pm 0.006 $ \\
\hline
$ 4.0 $ &$ 0.834 \pm 0.007 $ & $ 0.788 \pm 0.006 $ & $ 0.828 \pm 0.005 $ \\
\hline
$ 5.0 $ &$ 0.856 \pm 0.006 $ & $ 0.807 \pm 0.006 $ & $ 0.851 \pm 0.006 $ \\
\hline
$ 6.0 $ &$ 0.882 \pm 0.005 $ & $ 0.825 \pm 0.007 $ & $ 0.885 \pm 0.005 $ \\
\hline
		\end{tabular}}
  \end{subfloat}
  \caption{ (a) AUC scores associated with the ROC curves shown in
    Fig. \ref{fig:EP_vs_R1BCS_ROC_iid}, which correspond to EP, grVAMP and
    R1BCS based classification from i.i.d. Gaussian patterns in the
    presence of label noise, with $\eta=0.95$. (b) AUC scores associated 
    with the ROC curves shown in
    Fig. \ref{fig:EP_vs_R1BCS_ROC_cor}, which correspond to EP, grVAMP and
    R1BCS based classification from multivariate Gaussian patterns in the
    presence of label noise, with $\eta=0.95$.
    }
  \label{tab:EP_vs_R1BCS_iid}
\end{table}

\begin{table}
\begin{subfloat}[i.i.d. patterns \label{subtab:learned_eta_iid}]{
  \footnotesize
  \centering
  \begin{tabular}{|c|c|c|c|c|}
		\hline
		$\alpha$ & $\rho_L$ & $\Delta\rho/\rho$ & $\eta_L$ & $\Delta\eta/\eta$\\
		\hline
		2.5 & $ 0.229 \pm 0.004 $ & 0.08 & $ 0.964 \pm 0.001 $ & 0.02\\
		\hline
		3.0 & $ 0.234 \pm 0.003 $ & 0.07 & $ 0.957 \pm 0.003 $ & 0.007\\
		\hline
		4.0 & $ 0.247 \pm 0.004 $ & 0.01 & $ 0.9584 \pm 0.0005 $ & 0.009\\
		\hline
		5.0 & $ 0.249 \pm 0.003 $ & 0.003 & $ 0.9561 \pm 0.0007 $ & 0.006\\
		\hline
		6.0 & $ 0.252 \pm 0.003 $ & 0.007 & $ 0.9544 \pm 0.0004 $ & 0.005\\
		\hline
  \end{tabular}}
  \end{subfloat}
   \qquad
        \begin{subfloat}[patterns from MVN \label{subtab:learned_eta_cor}]{\adjustbox{valign=c}{
		\centering
		\footnotesize
  		\begin{tabular}{|c|c|c|c|c|}
		\hline
		$\alpha$ & $\rho_L$ & $\Delta\rho/\rho$ & $\eta_L$ & $\Delta\eta/\eta$\\
		\hline
		2.5 & $ 0.206 \pm 0.006 $ & 0.2 & $ 0.951 \pm 0.002 $ & 0.0006\\
		\hline
		3.0 & $ 0.208 \pm 0.006 $ & 0.2 & $ 0.953 \pm 0.001 $ & 0.003\\
		\hline
		4.0 & $ 0.228 \pm 0.005 $ & 0.09 & $ 0.953 \pm 0.001 $ & 0.003\\
		\hline
		5.0 & $ 0.23 \pm 0.005 $ & 0.08 & $ 0.9526 \pm 0.0005 $ & 0.003\\
		\hline
		6.0 & $ 0.236 \pm 0.004 $ & 0.05 & $ 0.9529 \pm 0.0004 $ & 0.003\\
		\hline
  \end{tabular}}}
  \end{subfloat}
  \caption{Values of the $\eta$
    parameter of the theta mixture pseudoprior estimated by the student 
    perceptron during the training phase when using EP to learn the weights of the teacher  (a) 
    from i.i.d. Gaussian patterns and (b) from multivariate Gaussian 
    patterns in the case $N=128$. The estimated value of $\eta$ is denoted as $\eta_L$, the true value being $\eta=0.95$, whereas the true  value of the density parameter of the spike-and-slab prior was given by $\rho=0.25$.
    }
\end{table}

One important limitation of the R1BCS algorithm as compared with EP (and grVAMP) is
that it involves both the inversion of a $N\times N$ matrix and of a
$M\times M$ matrix at each iteration. As a consequence, the
computational complexity of R1BCS is dominated by
$O((1+\alpha^3)N^3)$ operations. Therefore, from a computational
point of view, EP is especially advantageous as compared to R1BCS when the number of
patterns in the training set is large, as, in the EP formulation
proposed in this paper, the computational cost is of order $O((1+\alpha)N^3)$. However, in the large $N$ regime, grVAMP will be faster than EP (and  R1BCS) as the only operation that is $O(N^3)$ needs to be performed only once, whereas in EP the operations with cubic cost must be performed at each iteration. In fact, while the VAMP part in grVAMP is faster than EP for the sizes simulated in this paper, we recall that the grVAMP algorithm is composed of two modules (see also Fig. \ref{fig:grVAMP-FG}) implemented as two nested loops: the external loop corresponds to a minimum mean squared error estimation of the linear projection vector $\boldsymbol{z}=\mathbf{X}\boldsymbol{w}$, where $\mathbf{X}$ denotes the matrix of the patterns, under a Gaussian prior and a likelihood having the same functional form of the $\Lambda$ pseudoprior used in EP, whereas the inner loop consists of a VAMP module running on a standard linear model, whose measurement vector is given by the current estimate of $\boldsymbol{z}$, for a predefined number of iterations. We show the running times of R1BCS, EP and grVAMP in Table
\ref{tab:runtime_noisy} for the simulated size $N=128$. All the simulations were performed in parallel on a HP Proliant server with 64 cores clocked at 2.1GHz. In grVAMP, the maximum number of iterations of the outer MMSE module loop was set to 1000, while the number of iterations of the inner VAMP module was set to 2000. The comparisons shown in Table \ref{tab:runtime_noisy} are only meant to give an idea of the running times observed for the implementations that we used, which can be found at \texttt{https://github.com/abraunst/GaussianEP} for EP, \texttt{https://github.com/mengxiangming/glmcode} for grVAMP and \texttt{https://github.com/livey/R1BCS} for R1BCS. In particular, grVAMP was adapted by introducing the theta mixture measure and parameters were set in each algorithm as explained in this section. As the VAMP estimation is repeated for every iteration of the external loop, the running times related to grVAMP appear to be larger than those related to EP.

\begin{table}[h!tb]
    \begin{subfloat}[i.i.d. patterns \label{subtab:runtime_noisy_iid}]{\adjustbox{valign=c}{
		\centering
		\footnotesize
\begin{tabular}{|c|c|c|c|}
\hline
$\alpha$ & $t_{EP}$ (s) & $t_{grVAMP}$ (s)  & $t_{R1BCS}$  (s)\\
\hline
0.5 & $ 2.7 \pm 0.2 $ & $ 126.7 \pm 0.8 $ & $ 6.5 \pm 0.2 $ \\ 
\hline
1.0 & $ 2.6 \pm 0.03 $ & $ 134.6 \pm 0.9 $ & $ 12.0 \pm 0.3 $ \\ 
\hline
1.5 & $ 3.74 \pm 0.04 $ & $ 147.0 \pm 1.0 $ & $ 22.2 \pm 0.5 $ \\ 
\hline
2.0 & $ 5.65 \pm 0.08 $ & $ 159.0 \pm 1.0 $ & $ 47.6 \pm 1.0 $ \\ 
\hline
2.5 & $ 6.97 \pm 0.09 $ & $ 175.7 \pm 1.0 $ & $ 94.9 \pm 2.0 $ \\ 
\hline
3.0 & $ 8.3 \pm 0.1 $ & $ 212.9 \pm 20.0 $ & $ 150.9 \pm 3.0 $ \\ 
\hline
4.0 & $ 10.2 \pm 0.1 $ & $ 258.3 \pm 10.0 $ & $ 373.4 \pm 7.0 $ \\ 
\hline
5.0 & $ 12.2 \pm 0.1 $ & $ 311.6 \pm 10.0 $ & $ 628.0 \pm 10.0 $ \\ 
\hline
6.0 & $ 15.6 \pm 0.2 $ & $ 353.9 \pm 10.0 $ & $ 1052.9 \pm 20.0 $ \\ 
\hline
\end{tabular}}}
\end{subfloat}
\qquad
\begin{subfloat}[patterns from MVN \label{subtab:runtime_noisy_cor}]{\adjustbox{valign=c}{
		\centering
		\footnotesize
\begin{tabular}{|c|c|c|c|}
\hline
$\alpha$ & $t_{EP}$ (s) & $t_{grVAMP}$ (s) & $t_{R1BCS}$  (s)\\
\hline
0.5 & $ 2.7 \pm 0.2 $ & $ 139.2 \pm 2.0 $ & $ 6.7 \pm 0.2 $ \\ 
\hline
1.0 & $ 3.7 \pm 0.1 $ & $ 155.7 \pm 2.0 $ & $ 14.4 \pm 0.4 $ \\ 
\hline
1.5 & $ 4.8 \pm 0.1 $ & $ 206.8 \pm 20.0 $ & $ 27.0 \pm 0.7 $ \\ 
\hline
2.0 & $ 6.2 \pm 0.2 $ & $ 289.0 \pm 40.0 $ & $ 53.6 \pm 1.0 $ \\ 
\hline
2.5 & $ 7.7 \pm 0.2 $ & $ 236.3 \pm 20.0 $ & $ 105.8 \pm 3.0 $ \\ 
\hline
3.0 & $ 9.5 \pm 0.3 $ & $ 294.3 \pm 30.0 $ & $ 158.5 \pm 4.0 $ \\ 
\hline
4.0 & $ 11.4 \pm 0.3 $ & $ 366.9 \pm 40.0 $ & $ 379.4 \pm 8.0 $ \\ 
\hline
5.0 & $ 14.1 \pm 0.4 $ & $ 423.5 \pm 40.0 $ & $ 592.7 \pm 10.0 $ \\ 
\hline
6.0 & $ 15.9 \pm 0.4 $ & $ 498.5 \pm 50.0 $ & $ 1057.2 \pm 20.0 $ \\ 
\hline
\end{tabular}}}
\end{subfloat}

\caption{Running time related to the EP and R1BCS based sparse
  perceptron learning from (a) i.i.d. Gaussian patterns and from (b) Gaussian
  patterns from multivariate normal distribution with covariance
  matrix given by Eq. \eqref{eq: MVN covariance matrix} ($u=1$)
  in the presence of label noise, with $\eta=0.95$ and damping factor
  equal to 0.99 in the case of EP. The uncertainty on these values
  was estimated as $\sigma/\sqrt{N_{conv}}$, where $\sigma$ is the sample
  standard deviation over the set of converged trials and $N_{conv}$
  is the number of converged simulations.}
\label{tab:runtime_noisy}
\end{table}

\subsection{Correlated patterns generated by a recurrent neural network}
As an example of diluted network with correlated inputs, we consider a
network of $N$ randomly diluted perceptrons without self-loops. We
will denote the $i$-th row of the weight matrix $\mathbf{W}\in
\mathbb{R}^{N\times(N-1)}$ as $\boldsymbol{w}_i$. Each entry of
$\boldsymbol{w}_i$ is the weight of an incoming link of the $i$-th
perceptron. Each perceptron receives binary inputs $\boldsymbol{x}$
generated according to a Glauber dynamics at zero temperature. We
considered both the case of synchronous update of the patterns at each
time step and the case where the binary inputs are updated
asynchronously.

In the case of synchronous update, starting from an initial random
vector $\boldsymbol{x}_0=\text{sign}(\boldsymbol{\xi}_0)$, where
$\boldsymbol{\xi}_0\sim
\mathcal{N}(\boldsymbol{\xi};\boldsymbol{0},\mathbf{I})$ at discrete
time $t=0$ and given a pattern $\boldsymbol{x}^t$ at time $t$, each
perceptron computes its output at time $t+1$ according to:
\begin{equation}
z_{i}^{t}=\boldsymbol{w}_i^{T}\boldsymbol{x}_{\backslash i}^{t},
\label{eq:recnet_measurements}
\end{equation}
and to:
\begin{equation}
x_{i}^{t+1}=\text{sign}\left(z_{i}^{t}\right),
\label{eq:recnet_patterns}
\end{equation}
where $\boldsymbol{x}_{\backslash i}^{t}$ denotes the vector of
outputs produced by all perceptrons except the $i$-th one at time
$t$. The patterns at time $t+1$ are given by the set of outputs
resulting from Eq. \eqref{eq:recnet_measurements} and Eq.
\eqref{eq:recnet_patterns}.  While in principle such a recurrent
network dynamics could become trapped in a limit cycle when coupled to
a synchronous update rule for the patterns, in practice this never
happened when generating such patterns from a recurrent network of
$N=128$ diluted perceptrons, each of which having dilution level
$\rho=0.25$.

In the case of asynchronous updates, at each time step one perceptron
$i$ is selected at random and, given a pattern $\boldsymbol{x}^t$ at
time $t$, the $i$-th component of the pattern vector at time $t+1$ is
computed according to Eq. \eqref{eq:recnet_measurements} and Eq.
\eqref{eq:recnet_patterns}, while all other components are left
unchanged. In order to tune the degree of correlation of the patterns,
we generated them in such a way that two sets of patterns at
consecutive times have a given Hamming distance. In order to do so, we
ran the Glauber dynamics of the recurrent network as described and
updated the patterns asynchronously, but only stored their
configuration when the desired Hamming distance between the candidate
set of patterns at time $t+1$ and the set of patterns at time $t$ was
achieved.

We consider the noiseless case and analyze the performance of
training all the perceptrons of a student network of the same type of
the network that generated the patterns using both EP and grVAMP.  We
tested the two methods on a network with $N=128$ perceptrons and
density parameter of each perceptron given by $\rho=0.25$. We
considered both the case where the patterns are generated with the
synchronous update rule presented above and the case of asynchronous
update. In both cases, the damping parameter of EP was set to 0.999 and the convergence threshold was set to $10^{-4}$ for both EP and
grVAMP. In addition, the Gaussian part of the spike-and-slab prior was set to a standard normal distribution in both  EP and grVAMP. 

\begin{figure}
     \begin{subfloat}[]{
	 \includegraphics[width=0.47\textwidth]{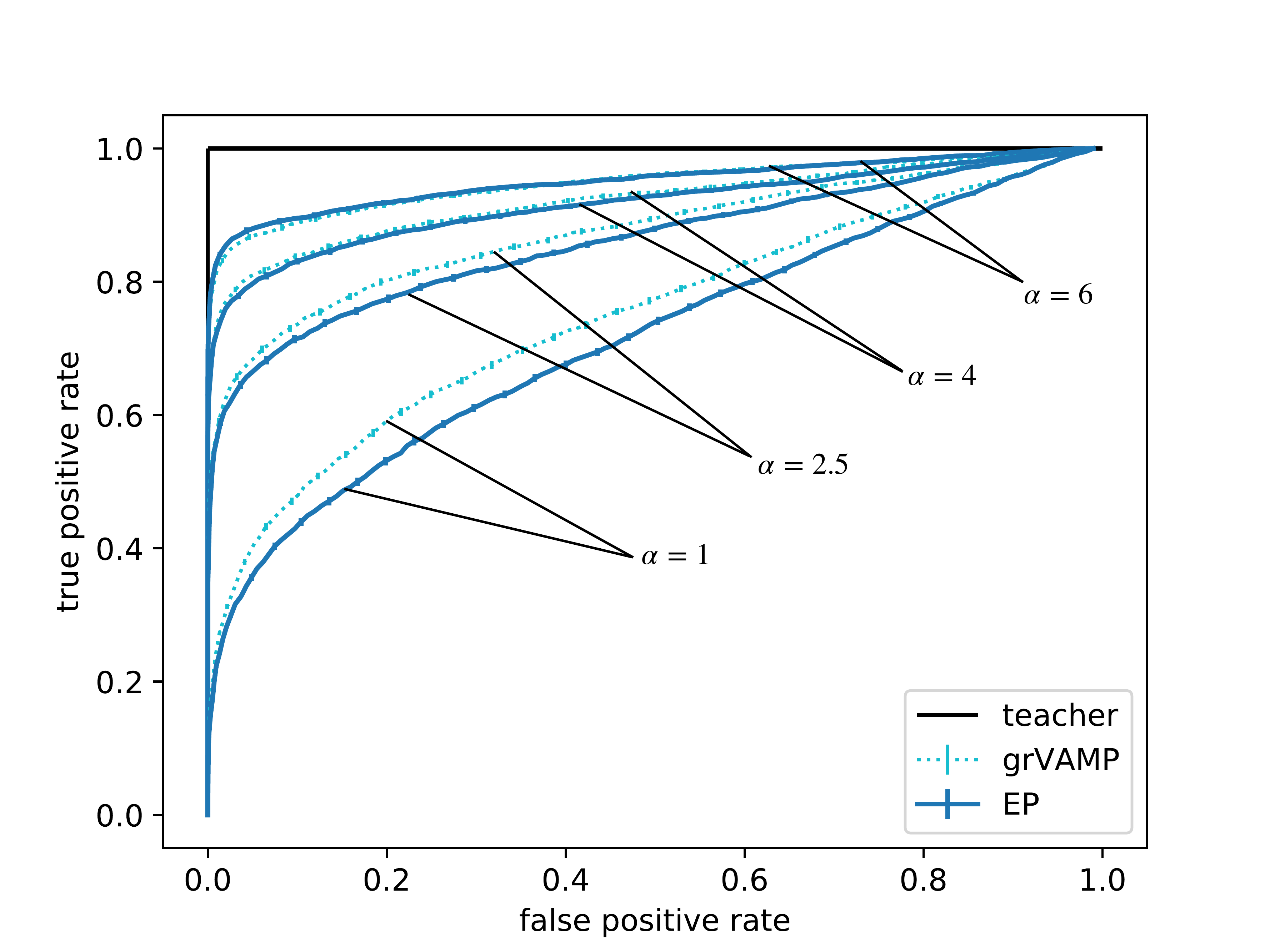}
      \label{fig:EP_vs_grVAMP_ROC_synchr}}
      \end{subfloat}      
      \hfill
      \begin{subfloat}[]{
      \includegraphics[width=0.47\textwidth]{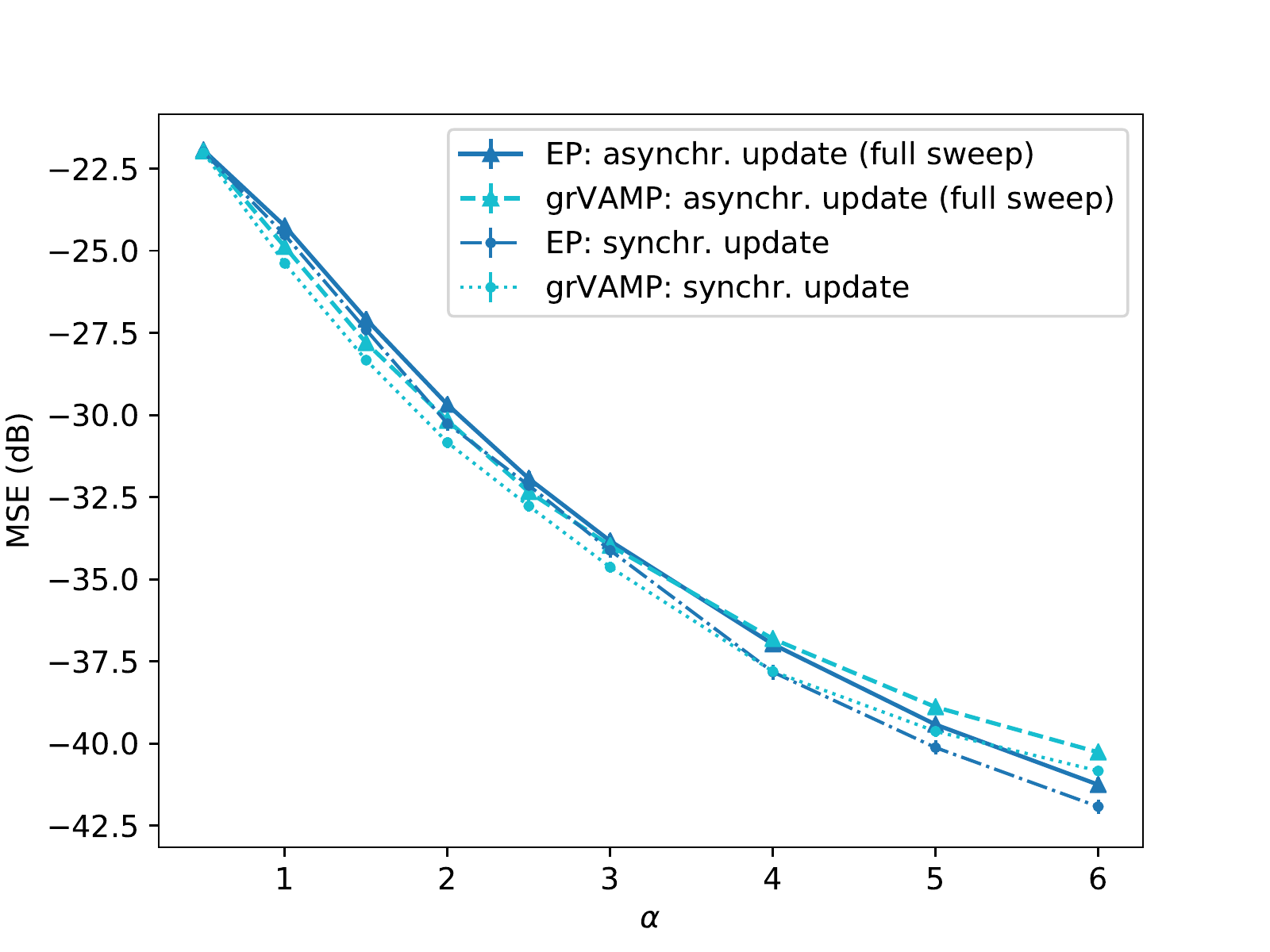}
      \label{fig:EP_vs_grVAMP_full_sweep}}
      \end{subfloat} 
      \\
      \begin{subfloat}[]{
      \includegraphics[width=0.47\textwidth]{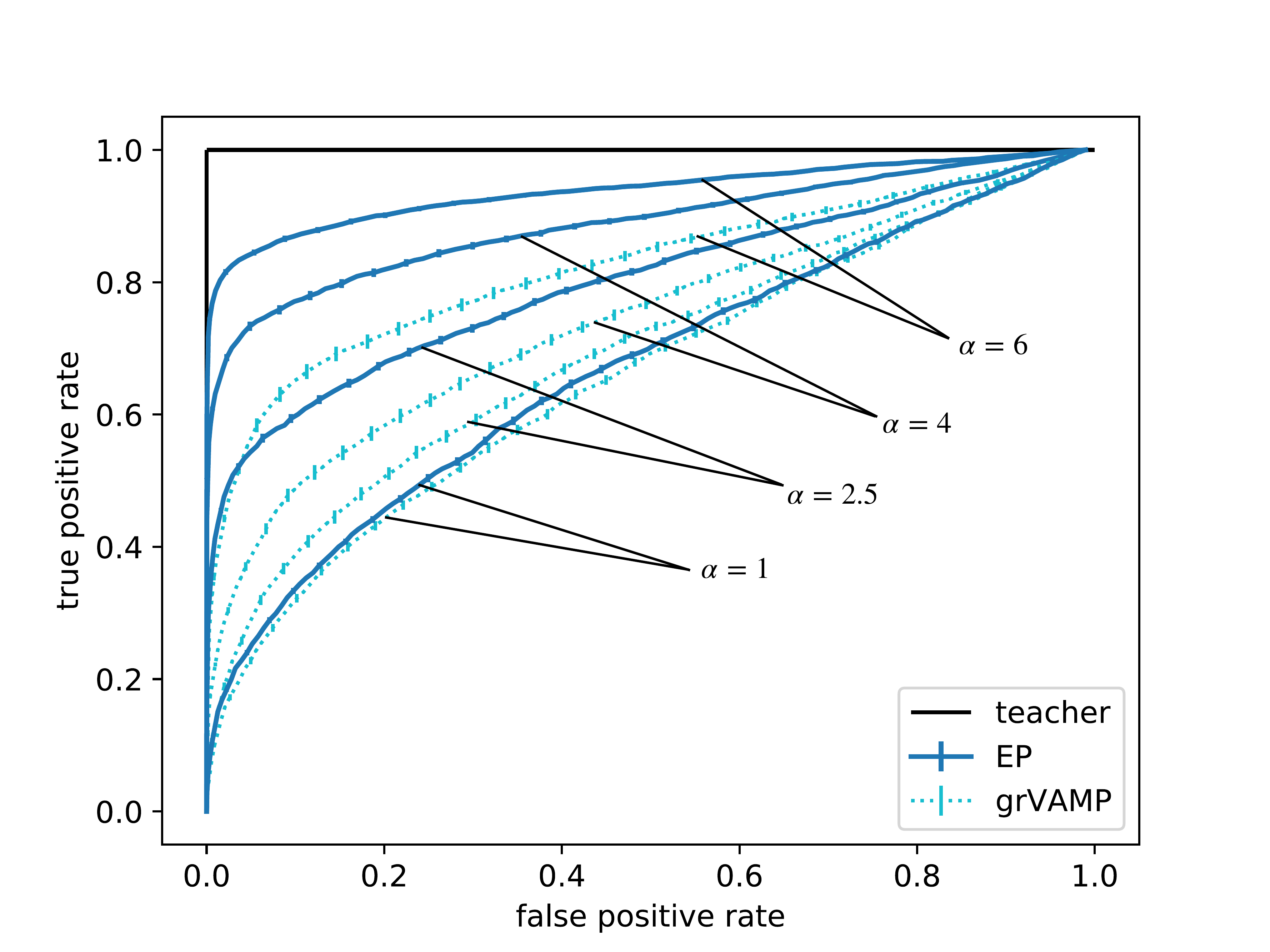}
      \label{fig:EP_vs_grVAMP_ROC_asynchr}}
      \end{subfloat}
      \hfill
 	\begin{subfloat}[]{
      \includegraphics[width=0.47\textwidth]{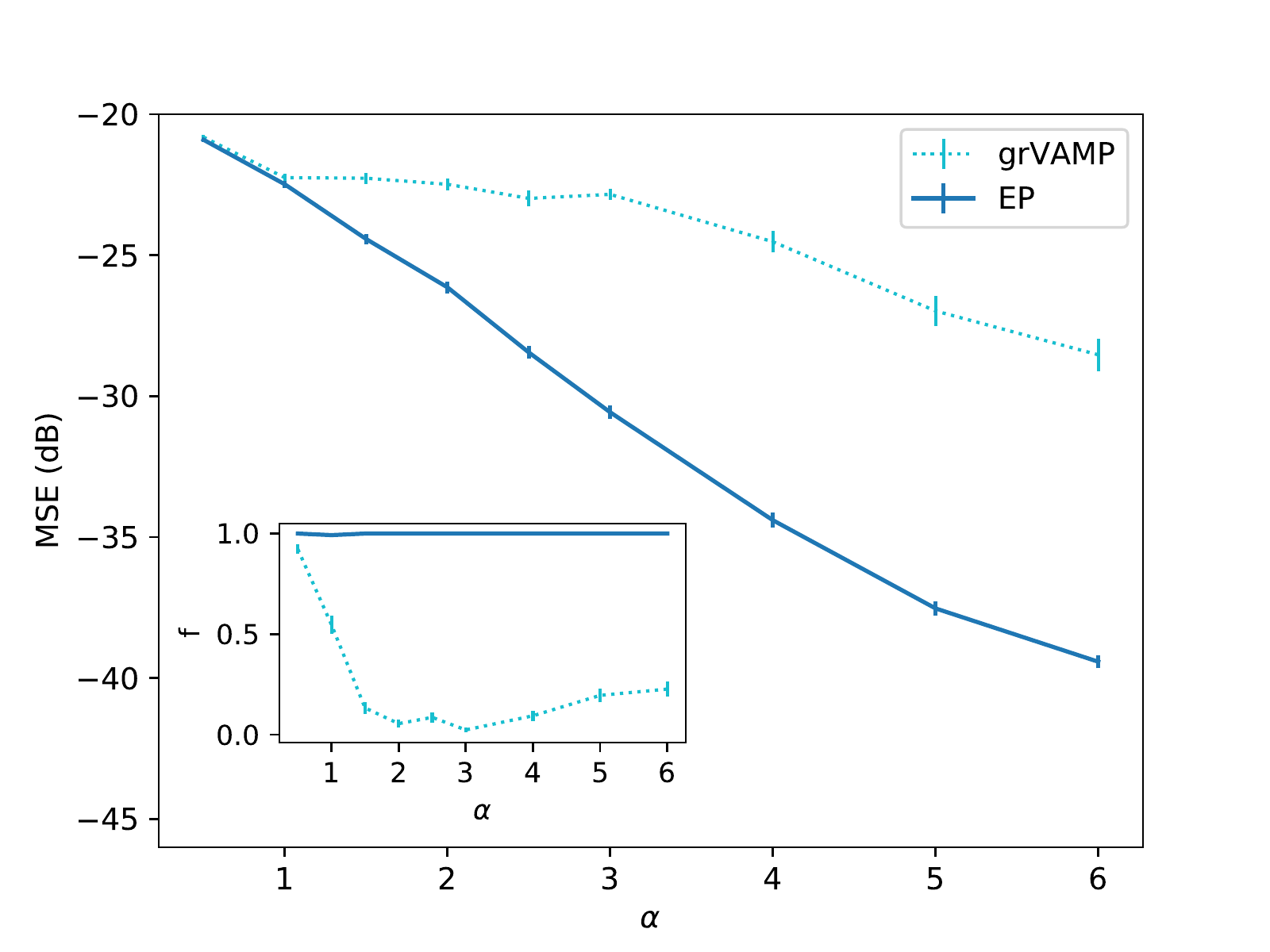}
      \label{fig:EP_vs_grVAMP_SensPlot_asynchr}}
      \end{subfloat} 
      \caption{Comparison between grVAMP and EP based 
        sparse perceptron learning from correlated patterns  
        generated by a recurrent network according to a Glauber
        dynamics. The case of \emph{weakly correlated} patterns and that of \emph{strongly correlated} patterns are shown in panels (a,b) and in panels (c,d), respectively. The number of perceptrons in the network is $N=128$ 
        and the density of the
        weights of each perceptron is $\rho=0.25$. In each plot, the mean values and the uncertainties were evaluated over the whole
        set of $N$ perceptrons. The error bars were estimated as $\sigma/\sqrt{N}$, 
        where $\sigma$ is the sample standard deviation computed over the set of 
        $N$ trained student perceptrons.
        (a) ROC curves related
        to the estimated topology
        of the learned weights of each perceptron for several values
        of $\alpha$ when patterns are synchronously updated. All $N$ perceptrons 
        achieved convergence during the training task. 
        (b)  MSE (in dB) associated with the 
        synchronously updated patterns of (a) and with the 
        asynchronous one in which patterns are included in the training set only after 
        each perceptron was selected to yield the corresponding update 
        (`full sweep'). (c) ROC curves related
        to the estimated topology
        of the weights learned from asynchronously updated patterns for several values
        of $\alpha$. The Hamming distance between 
        sets of patterns at consecutive times was chosen equal to 10.
       (d) MSE (in dB) corresponding to the case of the patterns considered in (c). The fraction of perceptrons of 
        the recurrent network whose training tasks 
        achieved convergence is shown in the inset.}
      \label{fig:EP_vs_grVAMP}
\end{figure}

When considering patterns generated with a synchronous
update, both algorithms achieved convergence during the training task
for all perceptrons of the recurrent network. We evaluated the ROC curves 
and the reconstruction error associated with the
correct/incorrect selection of the nonzero weights at the level of
single perceptron by considering the whole network and computing the
average and the standard deviation of the relevant quantities
(i.e. the MSE, the fraction of false positives and the fraction of
true positives) over all the perceptrons of the student network. Both algorithms
were able to select the same fraction of relevant weights, as shown by the
vertical portion of the corresponding ROC curves in Fig.
\ref{fig:EP_vs_grVAMP_ROC_synchr}, which were constructed by sorting
the weights according to their absolute value, in decreasing order. As a consequence, 
both grVAMP and EP showed comparable values for the MSE between the
weights estimated by the student perceptrons and the ones of the
corresponding teachers as a function of $\alpha$. Besides, similar values of the MSE were obtained for both algorithms by generating weakly correlated patterns according to an asynchronous generative process and including the generated vector of patterns in the training set only after a full sweep of $N$ updates, i.e. one for each perceptron of the teacher recurrent network. In this situation, the convergence rate of grVAMP was still very large ($\geq 95\%$) at all values of $\alpha$. We show the values of the MSE obtained from these two kinds of weakly correlated patterns in Fig. \ref{fig:EP_vs_grVAMP_full_sweep}.  

We repeated the same
numerical experiments with patterns generated performing an
asynchronous update of the perceptron outputs and fixing
$d_{H}(\boldsymbol{x}_{t+1},\boldsymbol{x}_{t})=10$, corresponding to
a Pearson correlation coefficient $r_{Pearson}=0.84$ between pattern
vectors at consecutive times. In this case, the observed convergence rate
within the student network was significantly lower for grVAMP than for
EP. Although the performance of EP and grVAMP were 
comparable when considering the subset of perceptrons whose grVAMP guided 
learning tasks converged (not shown), the ROC curves \ref{fig:EP_vs_grVAMP_ROC_asynchr} and MSE values \ref{fig:EP_vs_grVAMP_SensPlot_asynchr} obtained from these experiments, as evaluated over the $N$ perceptrons of the 
student network regardless of their convergence status at the end of the training task, show that grVAMP is not able to accurately estimate the weights of the teacher when it 
fails to converge. The fraction of perceptrons of the recurrent network whose
training task successfully converged is plotted in the inset of Fig. \ref{fig:EP_vs_grVAMP_SensPlot_asynchr} for both EP and grVAMP. These results strongly suggest that the observed convergence failure of grVAMP is related to the degree of correlation of the patterns presented to the student.

We verified that the student perceptrons were able to estimate the density of the 
weights of the teacher perceptrons quite accurately while learning the
classification rule of the teacher from this kind of patterns using EP and we checked that it is still possible to estimate the parameter
$\eta$ of the theta mixture pseudoprior during the training phase if a fraction of the labels are corrupted by noise, provided
that the noise level $1-\eta$ is not too large, similarly to what we
observed in the noisy case with i.i.d. patterns and with the
correlated patterns drawn from a multivariate normal distribution
analyzed in section \ref{subsec: noisy perceptron} of the paper. We
provide the estimated noise level in the noisy scenario where $\eta=0.9$ as well as some examples of the estimated values of the density at large
values of $\alpha$ in table \ref{tab:parameter_learning_recnet}. The estimate of the parameter $\eta$ proves to be fairly accurate in the case of patterns generated with a synchronous update rule, but is overestimated in the case where
the patterns are generated using an asynchronous update rule, as shown in table \ref{subtab:eta_learn_EP_recnet}.

\begin{table}
\begin{subfloat}[\label{subtab:eta_learn_EP_recnet}]{
		\centering
		\footnotesize
  		\begin{tabular}{|c|c|c|c|c|}
		\hline
		$\alpha$ & $\eta_L$, synchr. update & $\Delta\eta/\eta$, synchr. update & $\eta_L$, asynchr. update $(d_H=10)$ & $\Delta\eta/\eta$, asynchr. update\\
		\hline
		0.5 & $0.86 \pm 0.02$ & $0.05$ & $0.979 \pm 0.004$ & $0.09$\\
		\hline
		1.0 & $0.96 \pm 0.01$ & $0.06$ & $0.9945 \pm 0.0009$ & $0.1$\\
		\hline
		1.5 & $0.927 \pm 0.007$ & $0.03$ & $0.991 \pm 0.001$ & $0.1$\\
		\hline
		2.0 & $0.899 \pm 0.008$ & $0.001$ & $0.983\pm 0.001$ & $0.09$\\
		\hline
		2.5 & $0.893 \pm 0.007 $ & $0.008$ & $0.975\pm 0.001$ & $0.08$\\
		\hline
		3.0 & $0.901 \pm 0.001 $ & $0.001$ & $0.967 \pm 0.001$ & $0.07$\\
		\hline
		4.0 & $0.9036 \pm 0.0008$ & $0.004$ & $0.950 \pm 0.001$ & $0.06$\\
		\hline
		5.0 & $0.9087 \pm 0.0008$ & $0.01$ & $0.946 \pm 0.001$ & $0.05$\\
		\hline
		6.0 & $0.9108 \pm 0.0006$ & $0.01$ & $0.9389 \pm 0.0008$ & $0.04$\\
		\hline
\end{tabular}}
\end{subfloat}
\\
\begin{subfloat}[\label{fig:recurrent net learned density noiseless}]{
			\begin{tabular}{|c|c|c|}
			\hline
			Value of $\alpha$ & Estimated $\rho^\text{synchr}$ & Estimated $\rho^\text{asynchr},\ d_H=10$\\
			\hline
			$4.0$ & $0.224 \pm 0.002$ & $0.208 \pm 0.002$\\
			\hline
			$5.0$ & $0.233 \pm 0.002$ & $0.223 \pm 0.002$\\
			\hline
			$6.0$ & $0.237 \pm 0.001$ & $0.231 \pm 0.001$\\
			\hline
			\end{tabular}}
\end{subfloat}
\caption{(a) Estimate of the parameter $\eta$ of the theta mixture pseudoprior
  resulting from perceptron learning from the same patterns with label
  noise. The true unknown value of $\eta$ is 0.9. (b) Estimated value of the density of the weights of
        single perceptrons in the network at large alphas, both in the
        case of synchronously updated patterns and in the case of
        asynchronous update with fixed Hamming distance $d_H=10$
        between patterns at consecutive time steps, without label noise. In both (a) and (b),
  $N=128$ and $\rho=0.25$.}
\label{tab:parameter_learning_recnet}
\end{table}

\section{Conclusions}
\label{sec:conclusions}
In this paper we have proposed an expectation propagation based
strategy for efficient 1-bit compressed sensing reconstruction,
whose computational complexity is dominated by $O((1+\alpha)N^3)$ elementary operations. We analyzed the behavior in the zero
temperature case and assuming that the patterns are generated by a teacher having
the same structure as the student.

The performance of the algorithm has been extensively tested under
several conditions. For i.i.d. patterns generated by a
Gaussian distribution of zero mean and unit variance, the algorithm
performance are on par with two other state-of-the art algorithms:
1bitAMP \cite{onebitAMP} and grVAMP \cite{grVAMP}. However, in the
correlated Gaussian pattern case, where 1bitAMP fails to converge, EP outperforms 
grVAMP both in terms of accuracy and specificity. 
Moreover, both in the i.i.d. and correlated Gaussian case, EP is able to learn with 
remarkable accuracy the density $\rho$ of the weights of the teacher 
during the retrieval task. This feature of variational method (EP, 1bitAMP, grVAMP, etc.) puts them at an advantage over other
algorithms for 1-bit compressed sensing which require that the dilution level is 
known and provided among their inputs.

We then tested the robustness of EP reconstruction against noise. To
do so, we mislabeled a fraction $(1 - \eta )$ of the examples and compared EP with 
grVAMP and with the R1BCS algorithm
\cite{R1BCS} for 1-bit compressed sensing with sign-flip. Again, as
in the noiseless case discussed above, we first considered the i.i.d
pattern case and then the correlated Gaussian pattern case. 
The ROC curves and the sensitivity plots associated with EP, grVAMP and R1BCS
allow to conclude that the variable selection properties displayed by the 
three algorithms are not very different in the case of i.i.d. patterns. However, in the case of correlated Gaussian patterns and as long as the noise level affecting the labels is small enough, while on the one hand EP appears to be mostly comparable to R1BCS, on the other hand it systematically and consistently outperforms grVAMP regardless of the specific value of $\alpha$.  
In addition to the estimation of the density parameter, which we mentioned above in the case of sparse perceptron learning from noiseless examples, EP 
was able to successfully retrieve the noise level affecting the labels by including in the algorithm an analogous optimization strategy consisting in one step of gradient descent on the EP free energy at each iteration.
One important limitation of
R1BCS in comparison with EP, is the computational complexity: while 
the computational complexity of EP is $O((1+\alpha)N^3)$
(at least in the zero temperature case), R1BCS scales as
$O((1+\alpha^3)N^3)$ Therefore, from a computational point of
view, EP turns out to be especially advantageous in the large $\alpha$
regime as compared to R1BCS, but does not scale as well as grVAMP (and of course 1bitAMP) in terms of simulation time as a function of $N$.

Finally, we explored a more realistic scenario for ``temporally''
correlated patterns generated by a recurrent network of $N$ randomly
diluted perceptrons both in the case of synchronous and asynchronous
update schemes. First, we compared the performance of EP and grVAMP
in the noiseless case. The first striking observation is the poor
performance of grVAMP in terms of percentage of patterns for which the
iterative strategy converges. While EP converges basically on all the
presented sets of patterns, the convergence rate drops to less than 20\%
for a large interval of $\alpha$ values for grVAMP. EP in this regime
shows a remarkable performance both in terms accuracy and sensitivity and is still able to learn the parameters of the priors fairly well.

Taken together, these results show that EP is a competitive algorithmic
scheme with very good variable selection properties, particularly when 
one cannot rely on the statistical independence of the entries of the
pattern matrix. An important point of strength of
the algorithm is the possibility to infer {\it online} the optimal dilution of the
problem using a maximum likelihood data-driven iterative strategy, whereas, for other competitive strategies, the dilution is a
fixed parameter to set on the basis of some prior knowledge about the
problem. In addition, we have shown that the same maximum likelihood 
online strategy can be used to learn the consistency level of the labels when
they are affected by noise. As for the limitations of EP, the main disadvantage
is given by its cubic computational complexity, which makes it slower as 
compared with algorithms such as 1bitAMP and implementations of generalized VAMP. Furthermore, we found that, while EP was able to 
deal with the task of learning from noisy examples, a not too large level of noise 
was needed in order for the algorithm to learn the target classification rule with an 
acceptable accuracy. Finally, although in this work convergence was generally achieved within the thresholds we specified, it is well known that EP can suffer from numerical instabilities that prevent it from converging to its true fixed points. As a consequence, we expect that the results presented in this paper may not be the best results obtainable by means of the EP approximation. Attempting to better understand the convergence properties of EP in the settings presented in this paper as well as in other scenarios will be the object of future work.

\begin{acknowledgments}
AB, AP, and MP acknowledge funding from INFERNET, a European Union's
Horizon 2020 research and innovation program under the Marie
Sk\l{}odowska-Curie grant agreement No 734439, and from the SmartData@Polito
interdepartmental center for data science of Politecnico di Torino.
\end{acknowledgments}

\appendix

\section{Finite temperature formulation of Expectation Propagation}
\label{appendix: finite T EP}
We here recall the finite temperature EP scheme used in
\cite{braunstein2017} and state the update equations for the sparse
perceptron learning problem.

In this case, the posterior distribution is given by:
\begin{equation}
    P(\boldsymbol{h})=\frac{1}{Z_P}e^{-\frac{\beta}{2}\bm{h}^T\textbf{E}^{-1}\bm{h}}\prod_{i=1}^{N}\Gamma_i(h_i)\prod_{\tau=N+1}^{N+M}\Lambda_\tau(h_\tau),
    \label{eq:posterior_finitetemp}
\end{equation}
where we recall that:
\begin{equation}
\mathbf{E}^{-1}=\left(\begin{array}{cc}
\mathbf{X}_\sigma^{T}\mathbf{X}_\sigma & -\mathbf{X}_\sigma^T\\
-\mathbf{X}_\sigma & \mathbf{I}
\end{array}\right),
\end{equation}

As in section \ref{sec:methods}, we introduce Gaussian approximating
factors \eqref{eq:gaussianfactors} and write a fully Gaussian
approximation of the posterior distribution
\eqref{eq:posterior_finitetemp}:
\begin{equation}
\begin{split}
    Q(\boldsymbol{h}) &= \frac{1}{Z_Q}e^{-\frac{\beta}{2}\bm{h}^T\textbf{E}^{-1}\bm{h}}\prod_{i\in W}\phi(h_i;a_i,d_i)\prod_{\tau\in Y}\phi(h_\tau;a_\tau,d_\tau)=\\
    &=\frac{1}{Z_Q}\exp\left(-\frac{1}{2} (\bm{h} - \bar{\bm{h}})^T \bm \Sigma^{-1} (\bm{h} - \bar{\bm{h}})\right).
\end{split}
\label{eq:gaussposter_finitetemp}
\end{equation}
The covariance matrix $\bm{\Sigma}$ and the mean $\bar{\bm{h}}$ in
Eq. \eqref{eq:gaussposter_finitetemp} are given, respectively,
by:
\begin{equation}
    \boldsymbol\Sigma^{-1}=\beta\mathbf{E}^{-1}+\mathbf{D},
\end{equation}
and by:
\begin{equation}
\bar{\bm{h}}=\bm{\Sigma}\mathbf{D}\bm{a}.
\end{equation}
Moreover, we define the tilted distributions $Q^{(i)}$ for all $i=1,\dots,N+M$ as:
\begin{equation}
\begin{split}
    Q^{(i)}(\bm{h}) &= \frac{1}{Z_Q}e^{-\frac{\beta}{2}\bm{h}^T\textbf{E}^{-1}\bm{h}}\psi_i(h_i)\prod_{k\neq i}\phi(h_k;a_k,d_k)=\\
    &=\frac{1}{Z_{Q^{(i)}}} \psi_i(h_i)\exp\left(-\frac{1}{2} (\bm{h}-\bar{\bm{h}}^{(i)})^T \left(\bm{\Sigma}^{(i)}\right)^{-1} (\bm{h}-\bar{\bm{h}}^{(i)})\right),
    \end{split}
\end{equation}
where $\psi_i=\Gamma$ if $i\in W$ and $\psi_i=\Lambda$ if $i\in Y$.

As explained in section \ref{sec:methods}, the EP update equations for the means
$\bm{a}_i$ and for the variances $\bm{d}_i$ are obtained by imposing
the matching of the first and second moments of each variable $h_i$
w.r.t. the tilted marginal distributions $Q^{(i)}(h_i)$ and the fully
approximated marginal distributions $Q(h_i)$:
\begin{equation}
    \langle h_i \rangle_Q=\langle h_i \rangle_{Q^{(i)}},\qquad \langle h^2_i \rangle_Q=\langle h_i^2\rangle_{Q^{(i)}},
\end{equation}
which leads, again, to Eq. \eqref{eq:EP_update1} and to Eq. \eqref{eq:EP_update2}.

\section{EP free energy for the diluted perceptron problem}
Let us recall the definition of the EP free energy for a system with $N+M$ variables:

\begin{equation}
F_{EP}=(N+M-1)\log Z_{Q}-\sum_{k=1}^{N+M}\log Z_{Q^{(k)}},
\label{eq:EP_free_energy}
\end{equation}
where:
\begin{equation}
\log Z_{Q}=\frac{N+M}{2}\log\left(2\pi\right)+\frac{1}{2}\log(\det\bm{\Sigma}),
\end{equation}
and the expression of $\log Z_{Q^{(k)}}$ depends on the type of prior considered.
If $k=1,...,N$, the prior is a spike-and-slab and one has:
\begin{equation}
\log Z_{Q^{(k)}}=\log\left((1-\rho)\frac{1}{\sqrt{2\pi\Sigma_{k}}}e^{-\frac{\mu_{k}^{2}}{2\Sigma_{k}}}+\frac{\rho}{\sqrt{2\pi}}\sqrt{\frac{\lambda}{1+\lambda\Sigma_{k}}}e^{-\frac{1}{2}\frac{\lambda\mu_{k}^{2}}{1+\lambda\Sigma_{k}}}\right),\quad k=1,...,N
\label{eq:EP_free_energy_sparsity_contribution}
\end{equation}

For the remaining variables, the expression of $Z_{Q^{(k)}}$ either reads:
\begin{equation}
Z_{Q^{(k)}}=\sqrt{\frac{\pi\Sigma_{k}}{2}}\left(1+\text{erf}\left(\frac{\mu_{k}}{\sqrt{2\Sigma_{k}}}\right)\right),\quad k=N+1,...,N+M
\end{equation}
if $\Lambda(h_k)=\Theta(h_k)$, or it reads:
\begin{equation}
\begin{split}
Z_{Q^{(k)}}&=\sqrt{\frac{\pi\Sigma_{k}}{2}}\left[\frac{\eta}{2}\erfc\left(-\frac{\mu_k}{\sqrt{2\Sigma_k}}\right)+\frac{1-\eta}{2}\erfc\left(\frac{\mu_k}{\sqrt{2\Sigma_k}}\right)\right]=\\
&=\sqrt{\frac{\pi\Sigma_{k}}{2}}\left[\frac{1}{2}+\left(\eta-\frac{1}{2}\right)\erf\left(\frac{\mu_k}{\sqrt{2\Sigma_k}}\right)\right], \quad k=N+1,...,N+M,
\end{split}
\end{equation}
if $\Lambda(h_k)=\eta\Theta(h_k)+(1-\eta)\Theta(-h_k)$.

As a consequence, when $\Lambda(h_k)=\Theta(h_k)$, the EP free energy of the problem is given by:
\begin{equation}
\begin{split}
F_{EP}&=(N+M-1)\left(\frac{N+M}{2}\log\left(2\pi\right)+\frac{1}{2}\log(\det\bm{\Sigma})\right)-\\
&-\sum_{k=1}^{N}\log\left((1-\rho)\frac{1}{\sqrt{2\pi\Sigma_{k}}}e^{-\frac{\mu_{k}^{2}}{2\Sigma_{k}}}+\frac{\rho}{\sqrt{2\pi}}\sqrt{\frac{\lambda}{1+\lambda\Sigma_{k}}}e^{-\frac{1}{2}\frac{\lambda\mu_{k}^{2}}{1+\lambda\Sigma_{k}}}\right)\\
&-\frac{M}{2}\log\left(\pi/2\right)-\frac{1}{2}\sum_{k=N+1}^{N+M}\log\Sigma_{k}-\sum_{k=N+1}^{N+M}\log\left(1+\text{erf}\left(\frac{\mu_{k}}{\sqrt{2\Sigma_{k}}}\right)\right),
\end{split}
\end{equation}
whereas, when $\Lambda(h_k)=\eta\Theta(h_k)+(1-\eta)\Theta(-h_k)$, it is given by:
\begin{equation}
\begin{split}
F_{EP}&=(N+M-1)\left(\frac{N+M}{2}\log\left(2\pi\right)+\frac{1}{2}\log(\det\bm{\Sigma})\right)-\\
&-\sum_{k=1}^{N}\log\left((1-\rho)\frac{1}{\sqrt{2\pi\Sigma_{k}}}e^{-\frac{\mu_{k}^{2}}{2\Sigma_{k}}}+\frac{\rho}{\sqrt{2\pi}}\sqrt{\frac{\lambda}{1+\lambda\Sigma_{k}}}e^{-\frac{1}{2}\frac{\lambda\mu_{k}^{2}}{1+\lambda\Sigma_{k}}}\right)\\
&-\frac{M}{2}\log\left(\pi/2\right)-\frac{1}{2}\sum_{k=N+1}^{N+M}\log\Sigma_{k}-\sum_{k=N+1}^{N+M}\log\left(\frac{1}{2}+\left(\eta-\frac{1}{2}\right)\erf\left(\frac{\mu_k}{\sqrt{2\Sigma_k}}\right)\right).
\end{split}
\end{equation}

As a practical remark, an efficient way to compute $F_{EP}$
numerically involves using the Cholesky decomposition of the
covariance matrix in $\log(\det\bm{\Sigma})$, that is
$\bm{\Sigma}=\textbf{L}\textbf{L}^T$, where $\mathbf{L}$ is a lower
triangular matrix with real positive diagonal entries $L_{kk}$ for all
$k=1,\dots,N+M$. Then, the log determinant contribution to the EP free
energy is efficiently computed as
$\log(\det\bm{\Sigma})=2\sum_{k=1}^{N+M}\log(L_{kk})$.

\section{Learning the density level of the weights of the teacher and the noise on the labels}
\label{app:prior_param_learning}
The parameters of the prior distributions, such as the density $\rho$
of the weights of the teacher signal and the fraction $\eta$ of labels
fulfilling the consistency constraints, can be iteratively learned by
the student perceptron by minimizing the free energy associated with
the EP algorithm. We follow the reasoning laid out in reference
\citep{csep2020}, which we here recall and adapt to the sparse
perceptron learning problem.

Let $\boldsymbol\theta$ denote the set of parameters of the prior
distribution $P_{\boldsymbol{\theta}}(\boldsymbol{h})$. For example,
the density $\rho$ appears in the prior distribution in the factors
$\Gamma(w_i)$, for $i=1,\dots,N$, whereas the consistency level $\eta$
of the labels appears in the factors $\Lambda$ in the noisy case. Such
parameters can be estimated by the student perceptron by maximizing
the following likelihood function:
\begin{equation}
\begin{split}
P(\sigma_1,\dots,\sigma_M|\boldsymbol\theta,\boldsymbol{x}_1,\dots,\boldsymbol{x}_M)&=\int d \boldsymbol{h} P(\sigma_1,\dots,\sigma_M,\boldsymbol{h}|\boldsymbol\theta,\boldsymbol{x}_1,\dots,\boldsymbol{x}_M)=\int d \boldsymbol{h} P(\mathbf{X}_\sigma|\boldsymbol{h},\boldsymbol{x}_1,\dots,\boldsymbol{x}_M)P(\boldsymbol h|\boldsymbol\theta)=\\
&=\int d \boldsymbol{h} P(\mathbf{X}_\sigma|\boldsymbol{h},\boldsymbol{x}_1,\dots,\boldsymbol{x}_M) P(h_1,\dots,h_N|\rho,\lambda)P(h_{N+1},\dots,h_{N+M}|\eta)=Z(\boldsymbol\theta),
\end{split}
\label{eq:likelihood_params_sparsity_prior}
\end{equation}
which is nothing but the normalization of the posterior distribution in Eq. \eqref{eq:posterior}.

It is possible to associate a free energy to the partition function
\eqref{eq:likelihood_params_sparsity_prior} by using the definition
$F=-\log Z(\boldsymbol\theta)$.  When EP reaches its fixed point, $F$
is approximated by the EP free energy \eqref{eq:EP_free_energy} and
the student perceptron can attempt to minimize the latter via gradient
descent:
\begin{equation}
\theta_j^{(t+1)}=\theta_j^{(t)}-\delta\theta_j\frac{\partial F_{EP}}{\partial\theta_j},
\label{eq:gradient_descent_rho}
\end{equation}
where $t$ denotes the current iteration, $\theta_j$ denotes the $j$-th
component of the parameter vector $\boldsymbol\theta$ and
$\delta\theta_j$ is its corresponding learning rate, which was taken
equal to $10^{-5}$ in our numerical experiments both when learning
$\rho$ and when learning $\eta$.

Notice that, while the only contributions to $F_{EP}$ depending explicitly on the parameters $\boldsymbol\theta$ of the prior  are given by the terms $F_{Q^{(k)}}, k = 1,…,N+M$, the components of the gradient include other terms as well. However, these terms depend on the derivatives of the free energy with respect to the cavity parameters, which vanish at the EP fixed point. The optimization strategy we employ consists in iteratively alternating an EP update step at fixed prior parameters and an update of $\boldsymbol\theta$ performed via gradient descent at fixed EP parameters, similarly to an expectation maximization (EM) scheme, where the optimization over the EP parameters corresponds to the expectation step and the minimization of the free energy with respect to $\boldsymbol\theta$ corresponds to the maximization step. A ‘proper’ EM procedure would also be possible and would involve alternating a complete EP estimation of the approximating posterior distributions at fixed prior parameters until convergence is reached (E-step) and a maximum likelihood update of the prior parameters (M-step). The fact that we employ an alternating minimization procedure of this kind allows to only consider the explicit dependence of the free energy on the prior parameters.
In particular, we
have that $\partial F_{EP}/\partial\rho$ in Eq.
\eqref{eq:gradient_descent_rho} reads:

\begin{equation}
\frac{\partial F_{EP}}{\partial\rho}=\sum_{k=1}^{N} \frac{\frac{1}{\sqrt{2\pi\Sigma_{k}}}e^{-\frac{\mu_{k}^{2}}{2\Sigma_{k}}}-\frac{1}{\sqrt{2\pi(\lambda+\Sigma_{k})}}e^{-\frac{1}{2}\frac{\mu_{k}^{2}}{\lambda+\Sigma_{k}}}}{(1-\rho)\frac{1}{\sqrt{2\pi\Sigma_{k}}}e^{-\frac{\mu_{k}^{2}}{2\Sigma_{k}}}+\frac{\rho}{\sqrt{2\pi(\lambda+\Sigma_{k})}}e^{-\frac{1}{2}\frac{\mu_{k}^{2}}{\lambda+\Sigma_{k}}}}.
\end{equation}
Moreover, as highlighted in \citep{csep2020}, the second derivative
with respect to $\rho$ is strictly positive if $\lambda>0$, which
guarantees the uniqueness of the value of the density value that
minimizes $F_{EP}$, provided that the EP parameters $\mu_{k}$ and
$\Sigma_{k}$ are fixed for all $k=1,\dots,N$.

The same line of reasoning applies to the estimation of the parameter
$\eta$. In this case, the only contributions to $F_{EP}$ where $\eta$
appears are the terms $F_{Q^{(k)}},\ k=N+1,...,N+M$, and by taking the
derivative w.r.t. $\eta$ one obtains:
\begin{equation}
\frac{\partial{F_{EP}}}{\partial\eta}=\sum_{k=N+1}^{N+M}\frac{-2\text{erf\ensuremath{\left(\frac{\mu_{k}}{\sqrt{2\Sigma_{k}}}\right)}}}{1+(2\eta-1)\text{erf\ensuremath{\left(\frac{\mu_{k}}{\sqrt{2\Sigma_{k}}}\right)}}}.
\end{equation}

%
\end{document}